\documentclass[10pt,twocolumn]{article}

\usepackage{times}
\usepackage{epsfig}
\usepackage{graphicx}
\usepackage{amsmath}
\usepackage{amssymb}
\usepackage{caption}
\usepackage{subcaption}
\usepackage{algpseudocode}
\usepackage{algorithm}

\usepackage[pagebackref=true,breaklinks=true,,colorlinks,bookmarks=false]{hyperref}

\begin{document}

\title{EPiC: Ensemble of Partial Point Clouds for Robust Classification}

\author{Meir Yossef Levi, Guy Gilboa\\
Viterbi Faculty of Electrical and Computer Engineering \\
Technion - Israel Institute of Technology, Haifa, Israel\\
{\tt\small me.levi@campus.technion.ac.il ; guy.gilboa@ee.technion.ac.il}
}

\maketitle

\begin{abstract}
    Robust point cloud classification is crucial for real-world applications, 
    as consumer-type 3D sensors often yield partial and noisy data, degraded by various artifacts.
    In this work we propose a general ensemble framework, based on partial point cloud sampling. Each ensemble member is exposed to only partial input data. 
    Three sampling strategies are used jointly, two local ones, based on patches and curves, and a global one of random sampling.
    
    We demonstrate the robustness of our method to various local and global degradations.
    We show that our framework significantly improves the robustness of top classification netowrks by a large margin.
    Our experimental setting uses the recently introduced ModelNet-C database by Ren et al.\cite{modelnetc}, where we reach SOTA both on unaugmented and on augmented data. Our unaugmented  mean Corruption Error (mCE) is 0.64 (current SOTA is 0.86) and 0.50 for augmented data 
    (current SOTA is 0.57).
    We analyze and explain these remarkable results through diversity analysis. Our code is availabe at: \url{https://github.com/yossilevii100/EPiC}
\end{abstract}

\begin{figure}[t!]
    \centering
    \captionsetup[subfigure]{justification=centering}
    \begin{subfigure}[t]{0.1\textwidth} 
\includegraphics[width=1\textwidth]{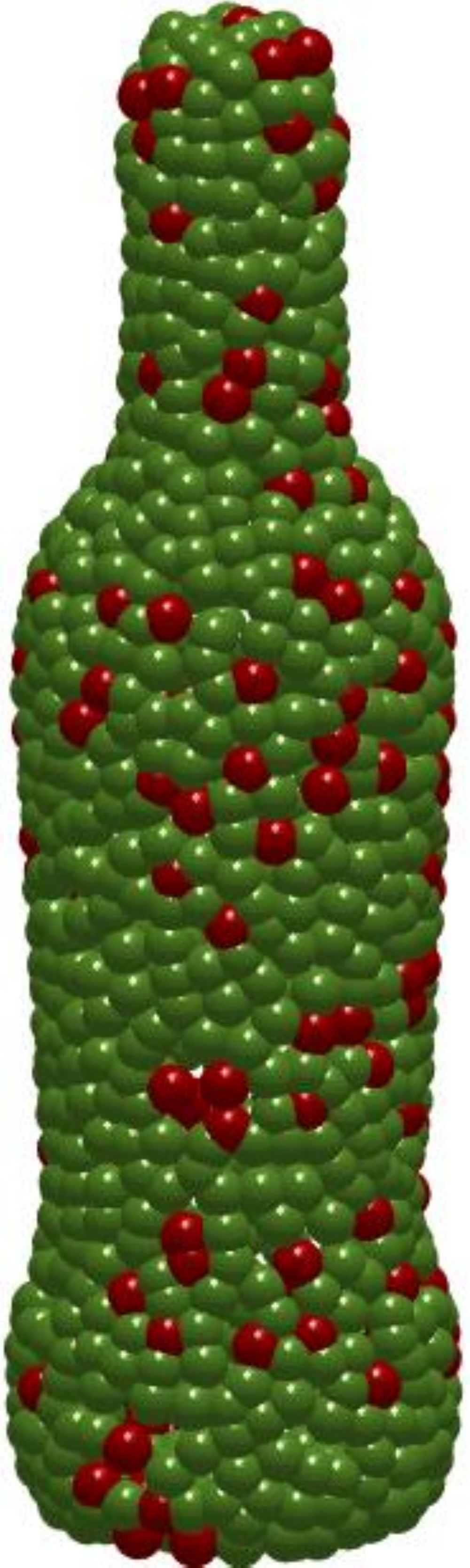}
\caption{Random}
        \label{subfig:random_sub_sampling}
    \end{subfigure}
    \begin{subfigure}[t]{0.1\textwidth} 
\includegraphics[width=1\textwidth]{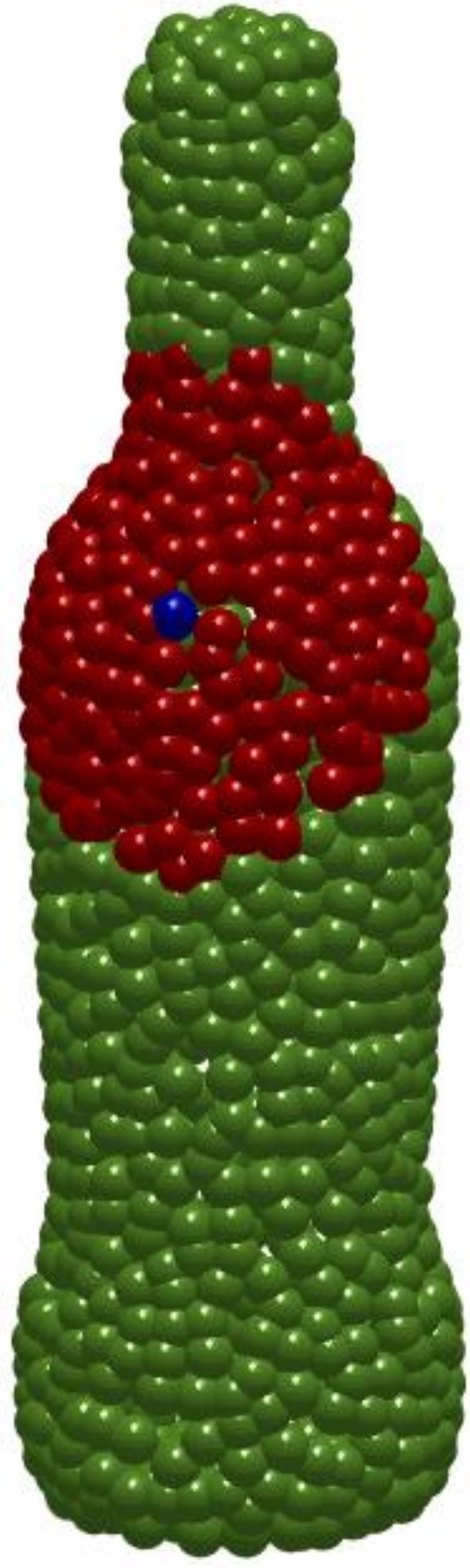}
\caption{Patch}    
        \label{subfig:patch_sub_sampling}
    \end{subfigure}
    \begin{subfigure}[t]{0.1\textwidth} 
\includegraphics[width=1\textwidth]{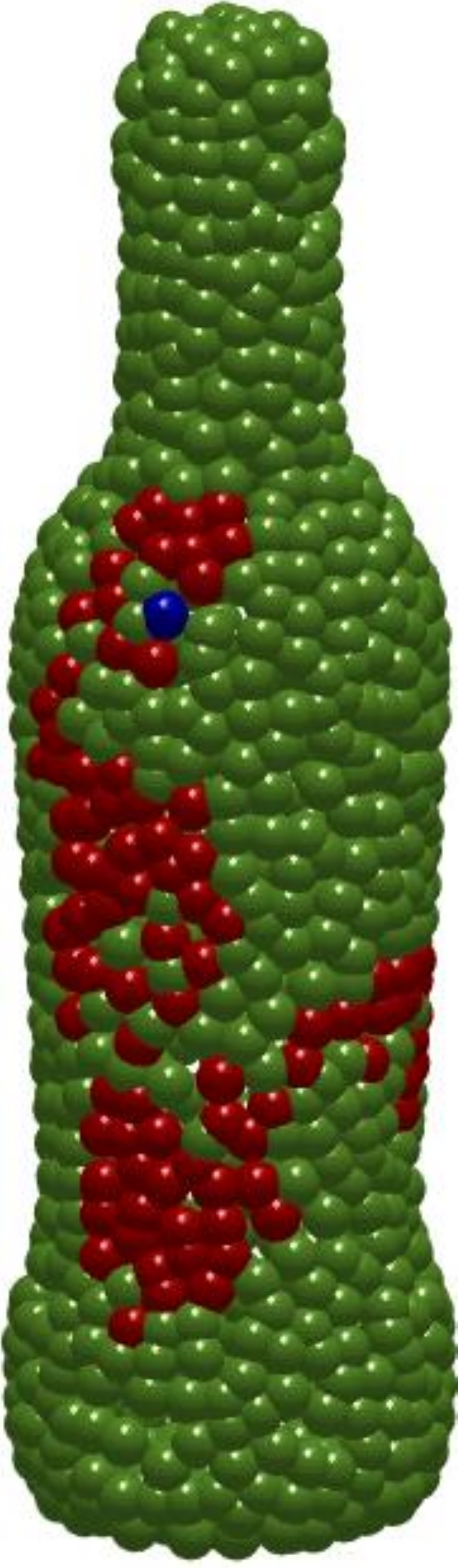}
\caption{Curve}
        \label{subfig:curve_sub_sampling}
    \end{subfigure}
    \caption{{\bf EPiC concept.} Three sampling mechanism are used in our ensemble: \textbf{Random} captures global information, \textbf{Patch} holds full local resolution, and  \textbf{Curve} is more exploratory in nature. {\bf Blue} - anchor point, {\bf Red} - sampled points. We show and explain why such ensembles are highly robust to various corruptions.
    }
    \label{fig:sampling}
\end{figure}
\section{Introduction}
\label{sec:intro}

A major obstacle in data-driven algorithms is the strong relation between performance and precise knowledge of input statistics. A way to test network robustness is to create corrupted test sets, where training is unaware of the specific corruptions. In recent years this has been investigated for images, with the creation of corrupted benchmarks e.g.: ImageNet-C, CIFAR10-C, CIFAR100-C \cite{imagenet_c} and MNIST-C \cite{mnist_c}.
In \cite{modelnetc} the idea is extended to point cloud classification, with the introduction of ModelNet-C. 
In our work we present a generic framework, based on sampling, which is light, flexible and robust to Out-Of-Distribution (OOD) samples.



Ensemble learning is a long-standing concept in robust machine learning \cite{adaboost, neural_network_ensembles, ensemble_methods}. We would like to obtain an ensemble of learners, which are loosely correlated, that generalize well also to OOD input.
There are two main questions: 1) How to form these learners?
   2) How to combine their results?

One classical way to form ensembles in deep learning (see for instance \cite{lakshminarayanan2017simple}), is to rely on the partial-randomness of the training process. In this setting, the ensemble consists of networks of the same architecture  trained on the same training-set. Variations of networks' output stem from the different random parameter initialization and the stochastic gradient descent process. This induces \emph{Stochastic diversity}.
Another major approach to generate ensembles (both in classical machine learning and in deep learning \cite{adaboost,lakshminarayanan2017simple,zefrehi2020imbalance}) is to change the sampling distribution of the training set for each learner. A mixture-of-experts approach is to 
train different types of classifiers, in neural networks this is accomplished by different network architectures. This induces \emph{Architecture diversity}. 
These approaches, however, may not yield sufficient OOD robustness (our experiments show their diversity is limited).
Deep ensembles are being investigated with new ways to calibrate and to estimate uncertainty through increasing the ensemble diversity \cite{zaidi2021neural, uncertainty3d}.

\begin{figure}[ptbh!]
  \centering
   \includegraphics[width = 1.1\linewidth, height =0.8 
   \linewidth]{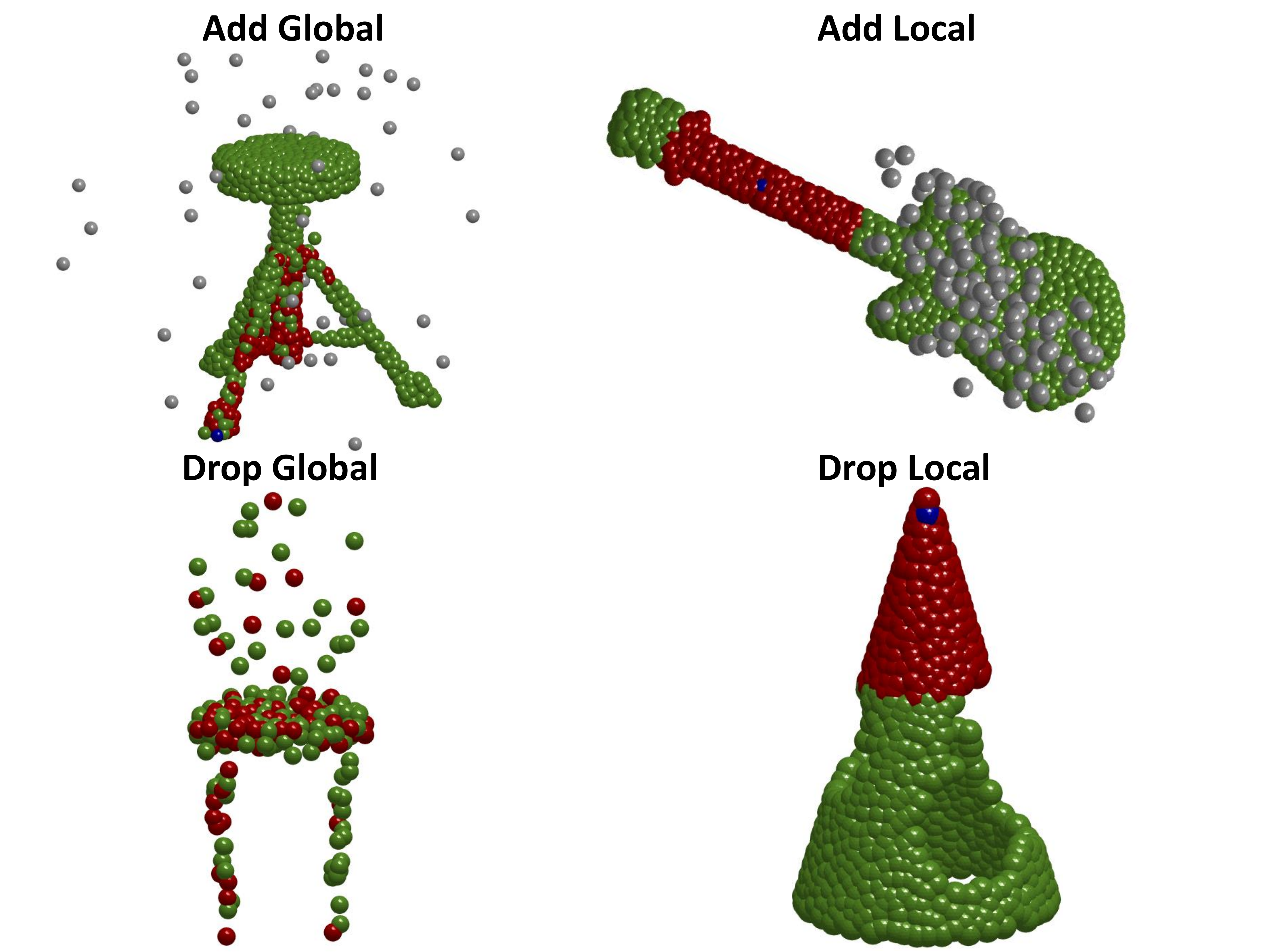}
   \caption{{\bf Overcoming corruptions by sampling}.
   Sampled partial point clouds (red points) have instances in which they are mostly not exposed to corruptions. The ensemble becomes highly robust to various unknown degradations. This is mostly apparent for non-uniform corruptions.
   }
   \label{fig:corruptions}
\end{figure}

Our approach to form ensembles is by exposing each ensemble member to limited input data. It is performed by generating different samples of each point cloud at both training and testing,
see Fig. \ref{fig:sampling}.
The ensemble becomes highly diverse, since each classifier has access to different parts of the data. Moreover, each sampling method has unique  characteristics. We observe that many types of corruption do not corrupt the data uniformly (see examples in Fig. \ref{fig:corruptions}). Some partial point clouds may thus be less corrupted, achieving accuracy closer to the case of clean data. Highly corrupted partial point clouds yield almost random network response, and can be modeled as adding noise to the classification decision. In this case, applying mean over the outputs of the ensemble significantly diminishes noise and improves accuracy . 
Four major advantages are gained by our proposed scheme:
1) The framework is completely generic and can essentially work with any point cloud classification network (we experimented with five, gaining improvement in all);
2) Many diverse classifiers can be generated;
3) Partial data is robust to local corruptions  and to outliers, performing well on OOD samples;
4) The required networks to be trained is the number of sampling methods (in our case three), and not the ensemble size.

\begin{table*}
  \centering
  \begin{tabular}{p{3.8cm} || p{1.0cm} p{1.0cm} p{1.0cm} p{1.0cm} p{1.1cm} p{1.1cm} p{1.0cm} p{1.0cm} p{0.8cm}}
    \hline
    Network (\#Ensemble size)& OA ↑ & mCE ↓ & Scale & Jitter & Drop-G & Drop-L & Add-G & Add-L & Rotate \\
    \hline
    DGCNN \cite{dgcnn} & 92.6\% & 1.000 & 1.000 &  1.000 & 1.000 & 1.000 & 1.000 & 1.000 & 1.000\\
    \hspace*{0.2cm} EPiC(\#12) (Ours) & 93.0\% & \underline{0.669} & 1.000 &  0.680 & 0.331 & 0.498 & 0.349 & \underline{0.807} & 1.019\\
    \hline
    GDANet\cite{gdanet} & 93.4\% & 0.892 & \textbf{0.830} & 0.839 & 0.794 &  0.894 & 0.871 & 1.036 &  0.981\\
    \hspace*{0.2cm} EPiC(\#12) (Ours) & 93.6\% & 0.704 & 0.936 &  0.864 & \underline{0.315} & 0.478 & 0.295 & 0.862 & 1.177\\
    \hline
    CurveNet\cite{curvenet} & \underline{93.8}\% & 0.927 & 0.872 & 0.725 & 0.710 &  1.024 & 1.346 & 1.000 &  \textbf{0.809}\\
    \hspace*{0.2cm} EPiC(\#12) (Ours)  & 92.1\% & 0.742 & 1.245 &  \textbf{0.617} & 0.363 & 0.585 & 0.495 & 1.029 & \underline{0.860}\\
    \hline
    PCT\cite{pct} & 93.0\% & 0.925 & 0.872 & 0.870 & 0.528 &  1.000 & 0.780 & 1.385 &  1.042\\
    \hspace*{0.2cm} EPiC(\#12) (Ours)  & 93.4\% & \textbf{0.646} & 0.894 &  0.851 & \textbf{0.306} & \textbf{0.435} & \underline{0.285} & \textbf{0.735} & 1.019\\
    \hline
    RPC\cite{modelnetc} & 93.0\% & 0.863 & \underline{0.840} & 0.892 & 0.492 &  0.797 & 0.929 & 1.011 &  1.079\\
    \hspace*{0.2cm} EPiC(\#12) (Ours)  & 93.6\% & 0.750 & 0.915 & 1.057 & 0.323 & \underline{0.440} & \textbf{0.281} & 0.902 & 1.330\\
    \hline
    PointNet\cite{pointnet} & 90.7\% & 1.422 & 1.266 &  \underline{0.642} & 0.500 & 1.072 & 2.980 & 1.593 & 1.902\\
    PointNet++\cite{pointnet++} & 93.0\% & 1.072 & 0.872 &  1.177 & 0.641 & 1.802 & 0.614 & 0.993 & 1.405\\
    RSCNN\cite{rscnn} & 92.3\% & 1.130 & 1.074 & 1.171 & 0.806 &  1.517 & 0.712 & 1.153 &  1.479\\
    SimpleView\cite{simple_view} & \textbf{93.9}\% & 1.047 & 0.872 & 0.715 & 1.242 &  1.357 & 0.983 & 0.844 &  1.316\\
    PAConv\cite{paconv} & 93.6\% & 1.104 & 0.904 & 1.465 & 1.000 &  1.005 & 1.085 & 1.298 &  0.967\\
    \hline
  \end{tabular}
  \caption{{\bf Main Experimental Result. ModelNet-C Unaugmented classification comparison.} {\bf Bold} best, \underline{underline} second best. Our framework dramatically improves robustness of all examined networks, as indicated by the mCE measure. Experiments using EPiC were conducted on the five most robust networks ($mCE \le 1$).  
  }
  \label{table:comparison_unaugmented}
\end{table*}


We reach robustness to various local and global degradations, yielding state-of-the-art results on corrupted ModelNet-C \cite{modelnetc} ($mCE = 0.646$ using PCT\cite{pct} and $mCE = 0.501$ using augmented with WolfMix\cite{pointwolf, rsmix} version of RPC \cite{modelnetc}), even with a small ensemble of size 12. 



\section{Related Work}
\label{sec:related}
{\bf Point Cloud classification.} It is customary to categorize point cloud classification networks to three mechanisms: multi-view, voxelizing, and point-wise networks. One of the challenges in point cloud processing is that, unlike images, 3D points are irregular in space. Multi-view methods project the point cloud into different viewpoints, generate 2D images, and apply CNN based networks \cite{multi_view_and_voxel, simple_view}. The projection forces the data to be regular. These approaches are slower because of the rendering phase, and might lose useful geometric information.
In the voxelizing mechanism, the solution for the irregularity problem is to  partition the 3D space into a voxel grid \cite{multi_view_and_voxel, modelnet40, voxel2}. This approach suffers heavily from sensitivity to the choice of grid regularity.
In the context of point-wise networks, PointNet \cite{pointnet} presented a pioneering method of applying MLP on the raw 3D points.
DGCNN\cite{dgcnn}, Dynamic Graph CNN, dynamically constructs a graph through the network. The graph is initially based on the raw 3D points, and progresses to more evolved  feature spaces with semantic connectivity. 
GDANet \cite{gdanet}, Geometry-Disentangled Attention Network, dynamically disentangles point clouds into contour and flat
components. Respective features are fused to provide distinct and complementary geometric information from each representation. Recently, PCT \cite{pct}, Point Cloud Transformer, adopted transformer architecture \cite{attention_is_all_you_need}
to create per-patch embedding, leveraging self attention mechanisms. CurveNet\cite{curvenet} generates features for each point by guided-walk over the cloud followed by a curve grouping operator.
Ren et al.\cite{modelnetc} studied the impact of several architectural choices on the robustness of the network and combined the most robust components to create RPC, which is the SOTA network on ModelNet-C. RPC takes advantage of 3D representations, using KNN, frequency grouping and self-attention mechanism that turned out to be the most robust combination. Another related work is Point-BERT\cite{pointbert} which splits the point cloud into patches as well. It then tokenizes the patches and uses pre-trained transformers with Mask Point Modeling. Point-MLP\cite{point_mlp} is based on residual MLPs, achieving impressive results on the clean data.


{\bf Robustness to shifts from the training set.} Traditionally, most of the attention in terms of robustness focused on basic deformations: jitter, scale and rotation.
There are different approaches in the literature focusing on some specific deformations, degradations or corruptions. For example, a long standing research topic is rotation invariance,
thoroughly addressed, e.g. by ClusterNet\cite{clusternet} and  LGR-Net\cite{lgr_net}.
PointCleanNet\cite{pointcleannet} and PointASNL\cite{point_asnl} addressed specifically robustness to outliers.
Studies investigating robustness to occlusions (common in 3D-sensors) use transformers \cite{IT_net} and voting mechanisms \cite{point_set_voting}.
Another aspect of robustness is related to 3D adversarial attacks, adressed by \cite{adversarial_pcs, extending_adversarial}. 
%
PointGuard\cite{pointguard} is also focused on guarding against adversarial attacks. $K$ random sampling procedures are performed for each point cloud, where the overall prediction is obtained by majority voting.
The theoretical analysis of \cite{pointguard} advocates the use of sampling very few points and using a huge ensemble ($K=10000$). 
Our experiments indicate that such ensembles do not perform as well on the clean data and are competitive only for certain types of OOD samples.

\begin{figure*}[ptbh!]
  \centering
   \includegraphics[width = 
   0.75\linewidth, height =0.38 
   \linewidth]{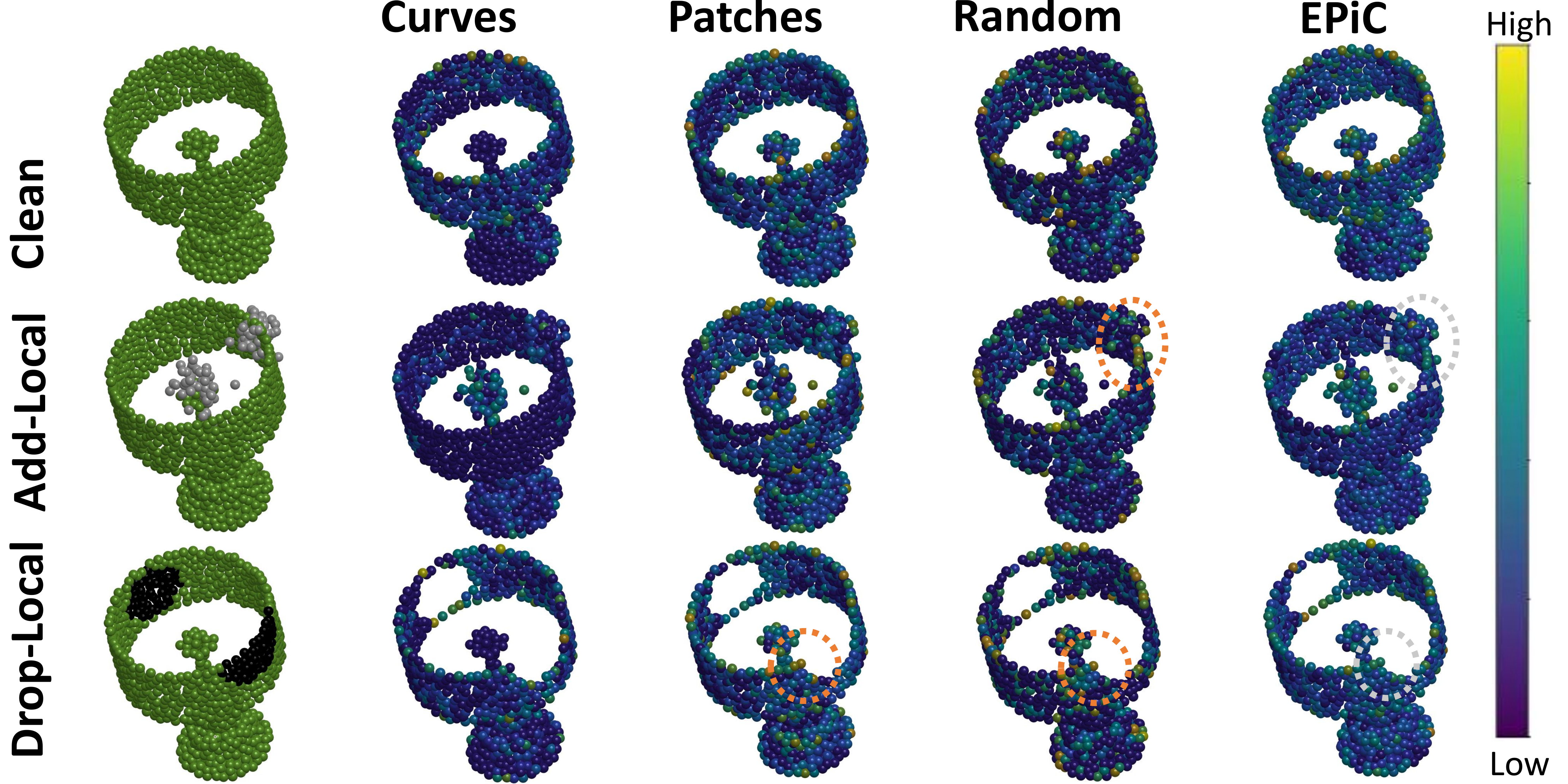}
   \caption{{\bf Pointwise Importance}. Left column - corruptions (grey - added points, black - removed points). The rest of the columns show color-coded pointwise importance. 
 Although Random is more influenced by Add-local, and Patches+Random by Drop-local (dashed orange circles), the aggregated result is robust to both (dashed grey circles).}
   \label{fig:intuition}
\end{figure*}

A major approach to increase robustness is by augmentation techniques which enrich the training set. RSmix\cite{rsmix} suggests mixing two samples from the database in a smooth manner, along with smoothing the labels of both samples, creating new virtual samples. This method inherently inserts jittering, scaling, translation and rotation. PointWolf\cite{pointwolf} applies non-rigid manipulations on selected anchor points. These include scaling, 3D rotation and translation with local jittering. Such complex manipulations are able, for example, to change a person's posture or the style of a chair. 
We note that augmentation introduces expected corruptions within the training set, thus violating, to some degree, the testing principles of OOD robustness.
To make our work complete, we examine the proposed framework also on augmented data (WolfMix\cite{modelnetc}, a combination of PointWolf and RSmix). We achieve SOTA in this case as well.

\subsection{Benchmarks}
\label{sec:benchmarks}
ModelNet-40 \cite{modelnet40} is a widely used dataset consisting of CAD meshes from 40 classes, such as airplane, chair and sofa. Each mesh was uniformly sampled to form a 3D point cloud consisting of 1024 points. The dataset has 12,311 samples, divided to 9843 for training and 2468 for test. The dataset contains closely semantic classes like vase, plant and flower pot, or chair and stool, which makes this dataset highly challenging.
Recently, in \cite{modelnetc} Ren et al proposed a corrupted point cloud benchmark to assess OOD robustness. It is based on the well studied ModelNet-40 and is referred to as ModelNet-C. 
We evaluate the robustness for OOD corruptions by applying it on ModelNet-C \cite{modelnetc}. This dataset contains seven types of corruptions: jitter, scale, rotate, add-global, add-local, drop-global and drop-local. Each of these corruptions is rendered in five levels of difficulty. In \cite{modelnetc} a unified calculation mechanism was defined to measure robustness. This measure is termed mCE (mean Corruption Error). It basically evaluates the results in comparison to DGCNN, which serves as a reference algorithm (and hence, by definition, has $mCE \equiv 1$). Each specific corruption has a similar score, relative to DGCNN, averaged over all five levels of difficulty. Since the measure is of error, a lower score is better. Please refer to 
\cite{modelnetc} for more details.



\section{Our Proposed Method}
\label{sec:method}

\subsection {Notations}
We denote by $N$ the number of cloud points at the input of a network (which, due to sampling, may vary). The number of features for each point in the cloud is $F$. We refer to $\tilde{K}$ as the ensemble size of each sub-sample mechanism and to $K$ as the combined ensemble size. In our case we suggest three sampling mechanisms of the same size, therefore $K = 3\tilde{K}$. $N_p$, $N_c$ and $N_r$ are the number of points in patches, curves and random sub-samples, respectively. In curve sampling there may be occasional repeating points, $N \le N_c$.
For curve extraction we have a hyper-parameter $M$ controlling the number of neighbors to choose from in a random-walk iteration. $C$ is the number of classes ($C=40$ in ModelNet-40), and $p \in \mathbb{R}^C$ is a single prediction vector. $P\in \mathbb{R}^{K \times C}$ is an ensemble of predictions. 

\subsection {Motivation of our approach}
The ability to classify in a robust manner is closely coupled with the ability to classify based on partial information. We would like a classification network to perform reasonably well also when some information is missing, noise is added or when outliers are introduced. Thus, it is desired to obtain a diverse set of features which are based on different locations in the shape. 
This could be demonstrated well in the experiment illustrated in Fig. \ref{fig:intuition}.

We visualize the internal classification importance of points for networks specializing in curves, patches and random. For the demonstration we focused on two degradations of adding and removing points locally. 
Commonly, a network gets a point cloud $X \in \mathbb{R}^{N\times 3}$ and encodes it into features $X_f \in \mathbb{R}^{N\times F}$ by a variety of sophisticated layers. The standard method to aggregate the points axis is by applying a symmetric function which is permutation invariant, such as max or mean.
In this example, in order to obtain features $X_f$, we use DGCNN\cite{dgcnn}.
We calculate the importance of each point $j$, denoted by $Imp(j)$, in the following manner,
\begin{equation}
    Imp(j) = \sum_{k=1}^F I(j==\arg\max_{n}(X_f(n,k))),
  \label{eq:importance}
\end{equation}
where  $I$ is an indicator function (1 when true and 0 otherwise).
The importance attempts to quantify the number of features of a specific point which are part of the global feature vector. This usually means that the feature's magnitude at that point is maximal, dominating the respective feature vector entries among all other points.
In this example, to cover the entire point cloud, we calculate the importance for all partial point clouds and present the average result in Fig. \ref{fig:intuition}.

When parts of the cloud are missing, or outliers are introduced, potentially corrupting features, we want the classification to be based on a variety of features, even though they may be less prominent in the clean set. Therefore, our aim is to ``spread'' the importance as evenly as possible. This insight motivates our approach for using partial point cloud ensembles to impose feature diversity. Additional analysis and insights appear hereafter.

\subsection{Proposed approach}
We use three types of sub-samples: two local ones, \emph{Curves} and \emph{Patches} and a global one consisting of \emph{Random-sampling}.
For the local methods, we first use farthest point sampling (FPS) algorithm \cite{pointnet++} to choose $\tilde{K}$ anchors. From each anchor we extract a patch and a curve. 

\textit {Patch extraction} is done by finding $N_p$ nearest neighbors. This sub-sample mechanism is more conservative, hence it preserves well the local information.

\textit {Curve extraction} is done by a random-walk process, beginning from the anchor point. $N_c$ random-walk iterations are performed, at each iteration one of $M$ nearest neighbors is chosen randomly. The choice is with replacement (hence the sampled partial point cloud may be smaller than $N_c$). This mechanism is more exploratory in nature and less structured.

\textit {Random extraction} is done by simply sub-sampling $N_r$ random points from the entire point cloud (without replacement, $N = N_r$).

The values of these parameters were determined once  and were used in the same manner in all our experiments for all classification networks (see details in Supp).

{\bf Generic framework.}
Our approach is generic and can be applied in conjunction with any point cloud classification network. 
The experiments (detailed in the experimental section) are conducted with five different architectures. 
These architectures are the most OOD-robust. Our method considerably improves the robustness of every one of them, as indicated by the mCE measure.
We use three instances of the same architecture. Each instance is trained to classify point clouds obtained by a certain sub-sampling mechanism. A recap of our approach (inference) is given in Algorithm \ref{alg:inference}, where the training procedure is detailed in the Supp.

\begin{algorithm}
\caption{Classification using EPiC (inference)}\label{alg:inference}
\begin{algorithmic}
\Require{$X, params_{Patches}, params_{Curves}, params_{Random}$} 
\State $model_{Patches} \gets params_{Patches}$
\State $model_{Curves} \gets params_{Curves}$
\State $model_{Random} \gets params_{Random}$
\State $anchors \gets FarthestPointSampling(X,\tilde{K})$

\For{$k \in \tilde{K}$} 
\State $Patch \gets FetchPatch(X, anchors(k))$\Comment{Local}
\State $Curve \gets FetchCurve(X, anchors(k))$\Comment{Local}
\State $Random \gets FetchRandom(X)$\Comment{Global}

\State $P_{Patch}^{k}\gets model_{Patches}(Patch)$
\State $P_{Curve}^{k} \gets model_{Curves}(Curve)$
\State $P_{Random}^{k} \gets model_{Random}(Random)$
\EndFor
\State $P_{ensemble} \gets Concatenate(P_{Patch}^{1:\tilde{K}}, P_{Curve}^{1:\tilde{K}}, P_{Random}^{1:\tilde{K}})$
\State $P \gets Mean(P_{ensemble})$; $Class=\arg\max(P).$

\end{algorithmic}
\end{algorithm}

\begin{table*}
  \centering
  \begin{tabular}{p{3.9cm} || p{1.0cm} p{1.0cm} p{1.0cm} p{1.0cm} p{1.1cm} p{1.1cm} p{1.1cm} p{1.0cm} p{0.8cm}}
    \hline
    Networks (\#Ensemble size)& OA ↑ & mCE ↓ & Scale & Jitter & Drop-G & Drop-L & Add-G & Add-L & Rotate \\
    \hline
    DGCNN\cite{dgcnn}+W.M & 93.2\% & 0.590 & 0.989 &  0.715 & 0.698 & 0.575 & 0.285 & 0.415 & \underline{0.451}\\
    \hspace*{0.2cm} EPiC(\#12)+W.M (Ours) & 92.1\% & 0.529 & 1.021 &  \textbf{0.541} & 0.355 & 0.488 & 0.288 & 0.407 & 0.600\\
    \hline
    GDANet\cite{gdanet}+W.M & \textbf{93.4}\% & 0.571 & \textbf{0.904} &  0.883 & 0.532 &  0.551 & 0.305 & 0.415 & \textbf{0.409}\\
    \hspace*{0.2cm} EPiC(\#12)+W.M (Ours) & 92.5\% & 0.530 & 0.968 & \underline{0.639} & 0.343 & 0.473 & 0.275 & 0.433 & 0.577\\
    \hline
    PCT\cite{pct}+W.M & \textbf{93.4}\% & 0.574 & 1.000 &  0.854 & 0.379 &  0.493 & 0.298 & 0.505 &  0.488\\
    \hspace*{0.2cm} EPiC(\#12)+W.M (Ours) & 92.7\% & \underline{0.510} & \underline{0.915} &  0.699 & \underline{0.323} & \underline{0.425} & \underline{0.268} & \underline{0.404} & 0.535\\
    \hline
    RPC\cite{modelnetc}+W.M & \underline{93.3}\% & 0.601 & 1.011 &  0.968 & 0.423 &  0.512 & 0.332 & 0.480 & 0.479\\
    \hspace*{0.2cm} EPiC(\#12)+W.M (Ours) & 92.7\% & \textbf{0.501} & \underline{0.915} &  0.680 & \textbf{0.315} & \textbf{0.420} & \textbf{0.251} & \textbf{0.382} & 0.544\\
    \hline
  \end{tabular}
  \caption{{\bf ModelNet-C Augmented 
  (WolfMix) classification Comparison.} {\bf Bold} best, \underline{underline} second best. 1) dramatically robustification gained by using EPiC on conventional methods. 2) Our method improves mCE and almost all OOD corruptions.}
  \label{table:comparison_augmented}
\end{table*}

\begin{table*}
  \centering
  \begin{tabular}{p{4.0cm} || p{0.85cm} p{0.95cm} p{0.75cm} p{0.8cm} p{1.1cm} p{1.1cm} p{1.0cm} p{1.0cm} p{0.75cm}}
    \hline
    Ensembles & OA↑ & mCE↓ & Scale & Jitter & Drop-G & Drop-L & Add-G & Add-L & Rotate \\
    \hline
    NS-1A \cite{dgcnn} & \underline{93.5}\% & 1.006 & \textbf{0.862} &  0.709 & 0.762 & 0.889 & 2.064 & 1.025 & \textbf{0.730}\\
    NS-3A \cite{dgcnn, pct, gdanet} & 93.4\% & 0.856 & \textbf{0.862} &  \underline{0.642} & 0.657 & 0.797 & 1.275 & 0.982 & \underline{0.777}\\
    \hline
    S-1A \cite{dgcnn}, Ours & 93.0\% & \textbf{0.669} & 1.000 &  0.680 & \underline{0.331} & \underline{0.498} & \underline{0.349} & \underline{0.807} & 1.019\\
    S-3A \cite{dgcnn, pct, gdanet}, Ours & \textbf{93.8}\% & \underline{0.671} &\underline{0.915} &  0.813 & \textbf{0.315} & \textbf{0.469} & \textbf{0.302} & 0.811 & 1.074\\
    \hline
    Point-Guard (\#1,000) \cite{pointguard} & 89.6\% & 0.949 & 1.947 &  \textbf{0.617} & 0.427 &  0.599 & 0.376 & \textbf{0.702} &  1.977\\
    \hline
  \end{tabular}
  \caption{{\bf ModelNet-C Ensemble methods classification Comparison.} {\bf Bold} best, \underline{underline} second best. Robustness improved by our suggested method.}
  \label{table:comparison_ensemble}
\end{table*}




\begin{figure}[ptbh!]
    \centering
    \captionsetup[subfigure]{justification=centering}
    \begin{subfigure}[t]{0.2\textwidth} 
\includegraphics[width=1\textwidth]{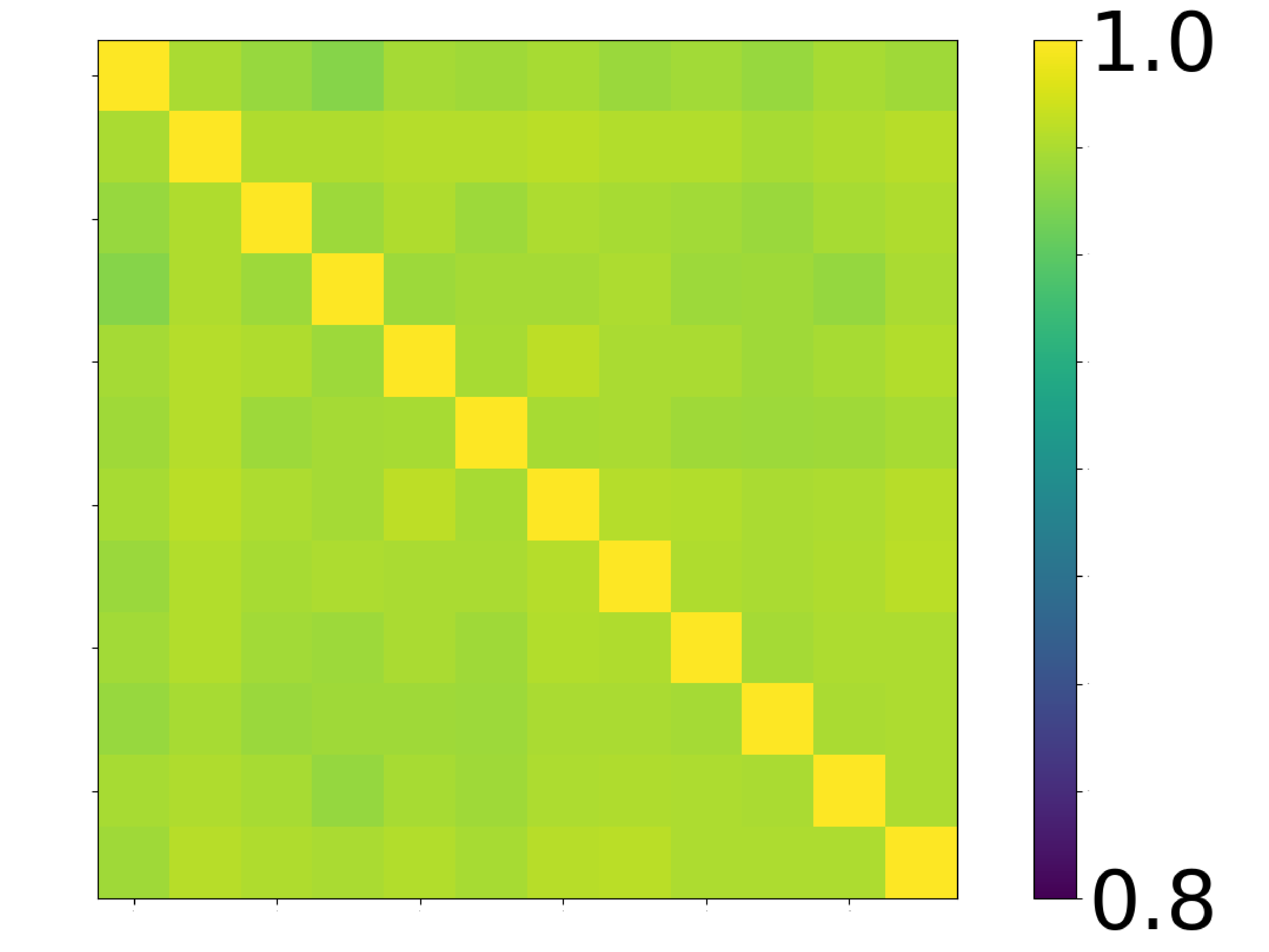}
\caption{{\bf NS-1A}, $c=0.948$}
        \label{subfig:clean_homo_homo}
    \end{subfigure}
    \begin{subfigure}[t]{0.2\textwidth} 
\includegraphics[width=1\textwidth]{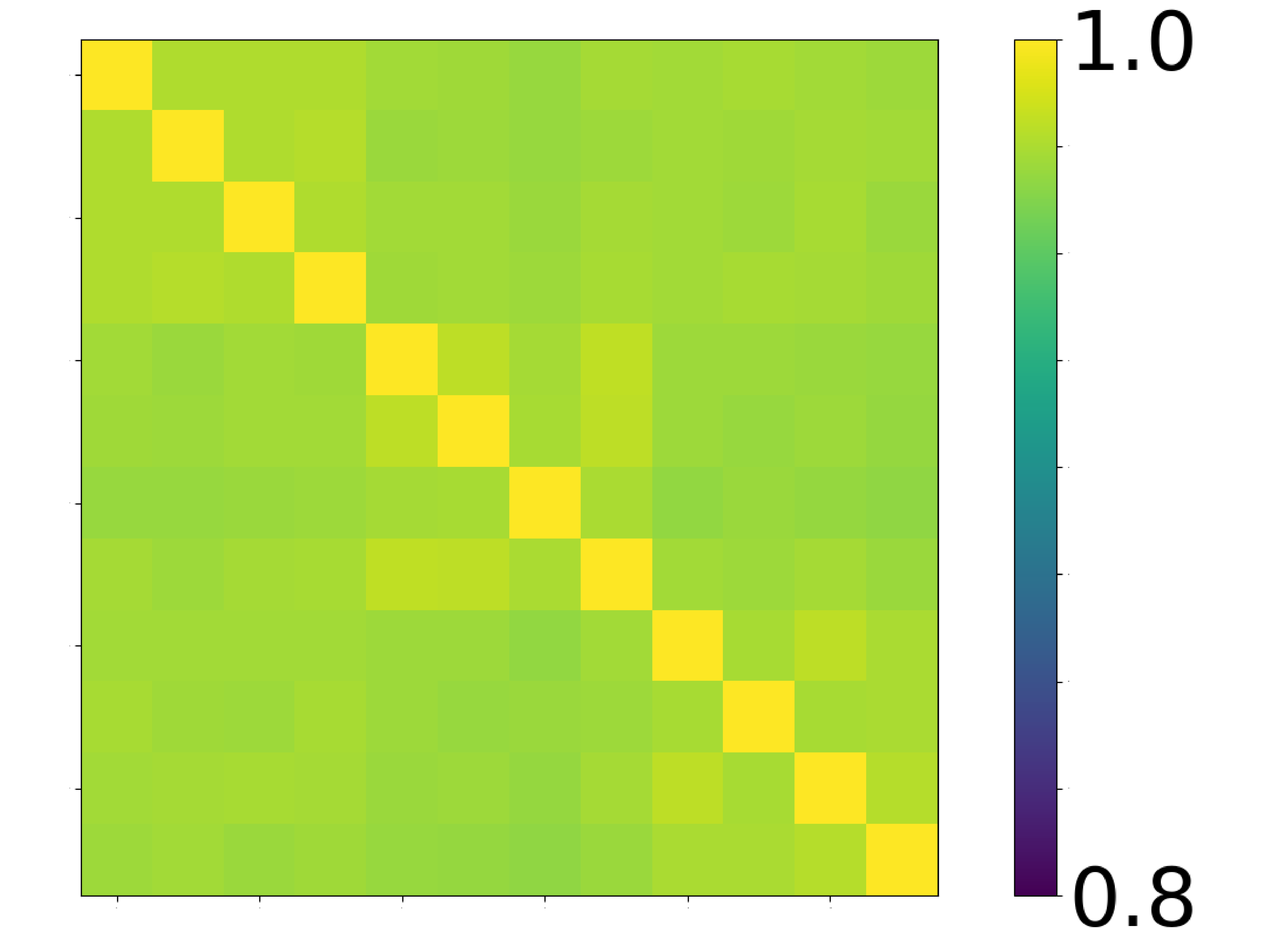}
\caption{{\bf NS-3A}, $c=0.940$}    
        \label{subfig:clean_homo_hetro}
    \end{subfigure}
    \begin{subfigure}[t]{0.2\textwidth} 
\includegraphics[width=1\textwidth]{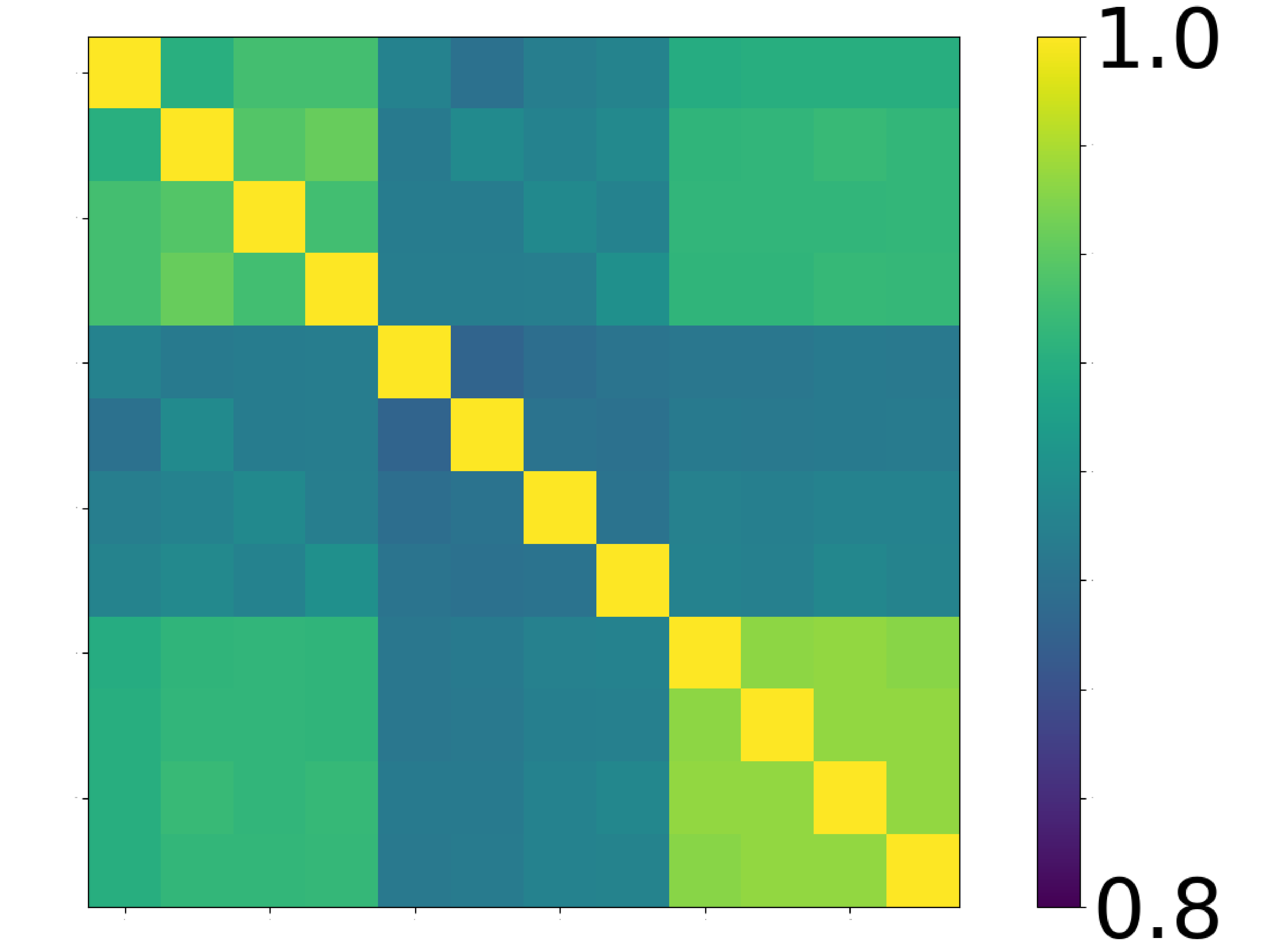}
\caption{{\bf S-1A}, $c=0.825$}
        \label{subfig:clean_hetro_homo}
    \end{subfigure}
    \begin{subfigure}[t]{0.2\textwidth} 
\includegraphics[width=1\textwidth]{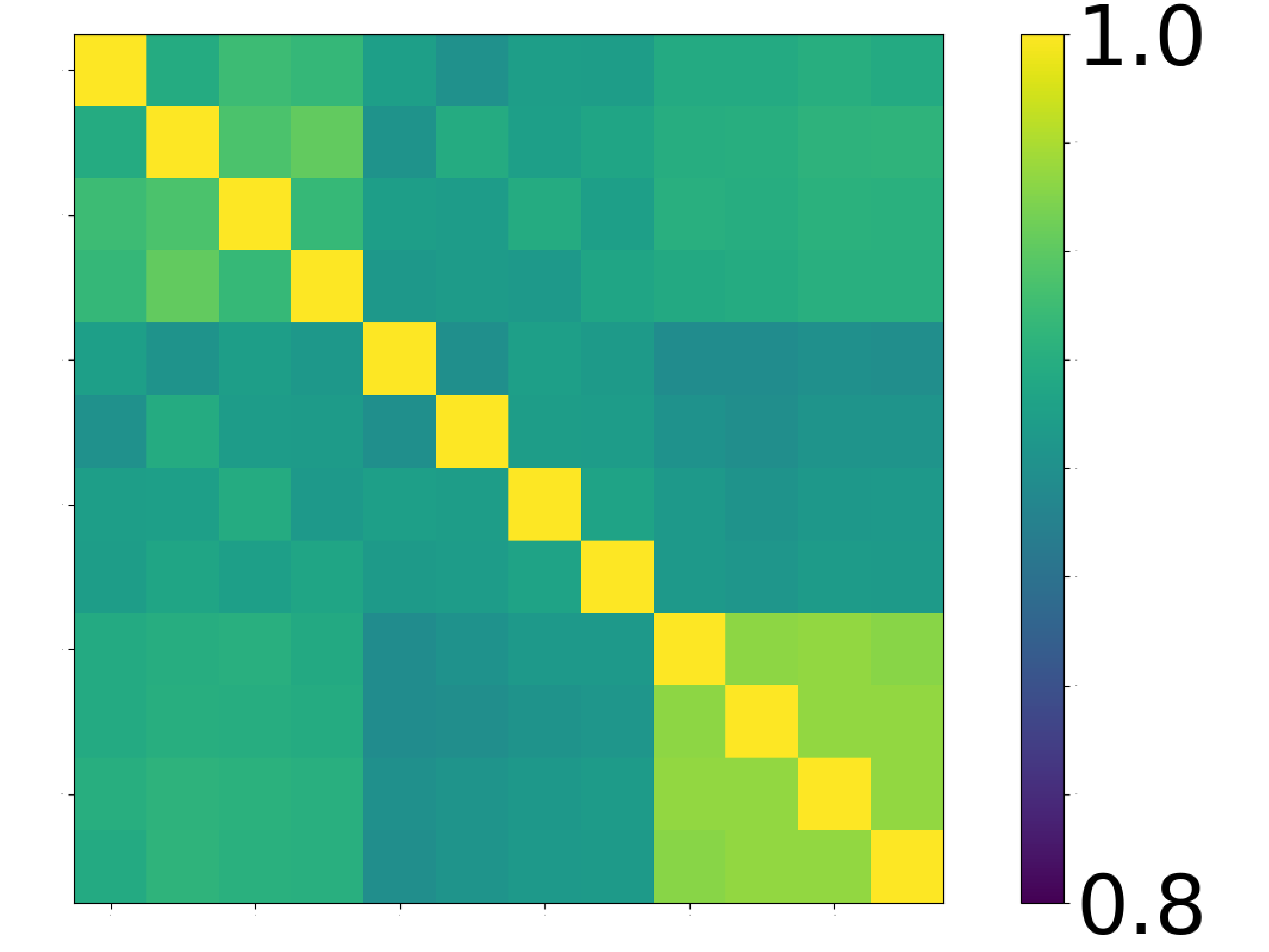}
\caption{{\bf S-3A}, $c=0.847$}
        \label{subfig:clean_hetro_hetro}
    \end{subfigure}
    \caption{\textbf{Correlation between ensemble members.} Correlation output of ensemble members on full non-sampled clean test-set (NS), top, with a single and three architectures, compared to sampling (S) by our approach, bottom, see setting details in Section \ref{sec:analysis}. $c$ is defined in Eq. \eqref{eq:C} (lower means higher diversity). Sampling affords higher diversity, compared to stochastic and architecture sources of diversity. 
    }
    \label{fig:corr}
\end{figure}

\section{Diversity Analysis}
\label{sec:analysis}
In order to leverage the advantage afforded by ensembles, a high level of diversity of ensemble members is required. 
This motivates us to quantitatively study the diversity of several types of ensembles. We investigate the following sources of diversity:
\begin{enumerate}
    \item {\bf Stochastic.} Diversity stemming from the stochasticity of the training process (initialization and SGD).
    \item {\bf Architecture.} Diversity caused by using different network architectures.
    \item {\bf Sampling.} Diversity due to different sampling methods and randomness of the sampling process.
\end{enumerate}
Four types of ensembles were examined (each consisting of 12 members, correlation results are shown in Fig. \ref{fig:corr}):
\begin{enumerate}
    \item {\bf No sampling, single architecture (NS-1A).} \emph{Stochastic diversity.} The ensemble consists of   12 instances of DGCNN\cite{dgcnn}. The entire point-cloud is used as input.
    \item {\bf No sampling, three architectures (NS-3A).} \emph{Architecture + stochastic diversity.}
    The ensemble consists of three architectures (each of 4 instances, in this order): PCT\cite{pct}, GDANet\cite{gdanet} and DGCNN\cite{dgcnn}. The entire point-cloud is used as input.
    \item {\bf Sampling (ours), single architecture (S-1A).} 
    \emph{Sampling diversity.}
    An ensemble consisting of three instances of DGCNN\cite{dgcnn}. Each instance is trained to specialize in a different sampling mechanism. The sampling methods (in this order) are: Patches, Curves and Random. Each instance uses 4 different sub-sample inputs.
    \item {\bf Sampling (ours), three architecture (S-3A).} 
    \emph{Sampling + architecture diversity.}
    Similar setting to the sampling diversity experiment. Here the following architectures, GDANet\cite{gdanet}, PCT\cite{pct} and DGCNN\cite{dgcnn}, were used on sampled inputs of Patches, Curves and Random, respectively. 
\end{enumerate}




In addition, PointGuard \cite{pointguard} was examined, as a different ensemble reference, with an ensemble of size $1000$.

\textbf{Measuring diversity through correlation.} 
We introduce a measure which quantifies (inverse) diversification by 
\begin{equation}
    c = \frac{1} {S}\sum_{i=1}^{S}{\frac{||C_i-I||^2} {K^{2}-K}},
  \label{eq:C}
\end{equation}
where $S$ is the dataset size, $K$ is the ensemble size, $I_{K\times K}$ is a unit matrix, $||\cdot||$ is the Frobenius norm and C is the Pearson correlation matrix of the ensemble predictions.
The final measure $c \in [0,1]$ is a scalar quantifying the diversity of the ensemble (in an inverse manner) for a given dataset.
For $c=0$ the members' response is completely uncorrelated (most diverse). For $c=1$ the members are fully correlated (zero diversity) and an ensemble is not necessary (would produce identical results as a single member). Note that since each member is quite accurate (a ``strong learner''), with an accuracy of around $90\%$, on binary problems we expect a diverse ensemble to be with $c \approx 0.9^2$ (in the multiclass case analysis requires additional assumptions, but should be in similar ranges).   
As can be seen in Fig. \ref{fig:corr}, ensembles based on sampling are considerably more diverse than those based on stochastic or architecture diversity. Moreover, curves and patches are lowly correlated, although the same anchor points are used. Diversity stems mostly from the sampling scheme, in addition to locality.
In Table \ref{table:comparison_ensemble} the improved robustness gained by the four ensemble methods is shown. 
Our partial-point-cloud strategy improves robustness in most criteria, excelling in mCE, with the three architecture configuration (S-3A) having superior results also in overall accuracy (clean set).





\section{Experiments}
\label{sec:exp}
We present our results for point cloud classification on ModelNet-C dataset, training on the clean dataset only and measuring performance on both clean and corrupted sets.
{\bf Implementation details.}
 We train the basic models independently on partial point clouds. 
 The predictions are aggregated using mean. For the unaugmented version we use only basic, standard augmentation procedures (detailed below) in order not to violate the OOD principle. For WolfMix augmented version we first augment the entire sample, then we generate the different sub-samples.
We eliminate the randomness with a fixed seed. All three models are trained simultaneously 300 epochs with learning rate of $5e^{-4}$, with cosine annealing scheduler \cite{cosine_annealing} to zero. We use a batch size of 256. For the unaugmented version we followed DGCNN \cite{dgcnn} protocol for augmentation: 1) random anisotropic scaling in the range $[2/3, 3/2]$; 2) random translation in the range $[-0.2, +0.2]$. The implementation uses Pytorch library \cite{pytorch}. Cross-Entropy loss is minimized.
In the training phase we split each sample to 4 farthest point samples, and independently predict each partial point cloud class. The inference procedure is detailed in Algorithm \ref{alg:inference}.
\begin{table*}[ht]
  \centering
    \begin{tabular}{p{4cm} || p{1.0cm} p{1.0cm} p{1.0cm} p{1.0cm} p{1.1cm} p{1.1cm} p{1.0cm} p{1.0cm} p{0.8cm}}
    \hline
    Sub-samples & OA ↑ & mCE ↓ & Scale & Jitter & Drop-G & Drop-L & Add-G & Add-L & Rotate \\
    \hline
    DGCNN-Curves (\#4) & 90.7\% & 1.069 & 1.628 & 1.297 & 0.431 & 0.729 & \underline{0.363} & 1.618 & 1.414\\
    DGCNN-Patches (\#4) & 92.8\% & 0.793 & \textbf{0.989} & 1.165 & 0.577 & 0.536 & 0.505 & 0.851 & \textbf{0.930}\\
    DGCNN-Random (\#4) & 91.5\% & 0.766 & 1.234 & \textbf{0.399} & \underline{0.351} & 0.812 & 0.580 & \textbf{0.793} & 1.195\\
    \hline
    DGCNN-Mean (\#12) & \textbf{93.0}\% & \textbf{0.669} & \underline{1.000} &  \underline{0.680} & \textbf{0.331} &  \textbf{0.498} & \textbf{0.349} & 0.807 &  \underline{1.019}\\
    DGCNN-Maj. Voting (\#12) & \underline{92.6}\% & \underline{0.706} & 1.043 &  0.794 & 0.359 &  \underline{0.517} & 0.380 & \underline{0.800} &  1.051\\
    
    \hline
  \end{tabular}
  \caption{\textbf{Sub-Samples vs. Aggregated.} \textbf{Bold} best among aggregations, \underline{underline} best among sub-samples. Aggregations are almost
always superior for any corruption. Thus, it can be inferred that the partial classifiers are low-correlated.}
  \label{table:aggregations}
\end{table*}

\subsection{Main evaluation results}
The EPiC framework is implemented using several widely used point cloud classification networks.
Table \ref{table:comparison_unaugmented} shows the  Unaugmented results and Table \ref{table:comparison_augmented} shows results following augmentation by WolfMix \cite{pointwolf, rsmix}.

\textbf{Unaugmented.} A major advantage is achieved in terms of mCE for all networks. Using PCT\cite{pct} with EPiC reaches mCE=0.646, which outperform current SOTA of RPC\cite{modelnetc} (mCE=0.863) by a large margin. With respect to accuracy on the clean data set (OA), we see improvement in four out of five cases, with only CurveNet\cite{curvenet} degrading, where RPC\cite{modelnetc} and GDANet\cite{gdanet} reach 93.6\%.

\textbf{Augmented (WolfMix).} We further examined wheather our method can improve augmentation procedures (which to some extent violate some OOD assumptions). EPiC on augmented data consistently improves robustness. Using RPC \cite{modelnetc} EPiC achieves mCE=0.501 surpassing current augmented SOTA (mCE=0.571). OOD robustness comes at a cost of minor accuracy drops, consistently for all networks. The trade-off between accuracy and robustness is well demonstrated in \cite{modelnetc} for point-cloud classification, and in \cite{robustness} for general classification settings.

\subsection{When our method performs best?} 

Let $X \in \mathbb{R}^{N\times 3}$ be the set of coordinates of a clean point cloud. Following a corruption transformation $T_c$ acting on $X$ we get a corrupted point cloud $X_c \in \mathbb{R}^{M\times 3}$, where $X_c=T_c(X)$. Let the intersection of these sets be define by $X_{\cap}:= X \cap X_c$ of size $|X_{\cap}|$ points. We define the \emph{uniformity} of the corruption transformation by
\begin{equation}
    \label{eq:u}
    u(X,T_c) := 1 - \frac{|X_{\cap}|}{max(N,M)}.
\end{equation}
For a fully uniform corruption $T_c$, all points of the original point cloud change, hence $X_{\cap}=\emptyset$ and $|X_{\cap}|=0$, yielding $u=1$. When the transformation is highly selective, affecting only a few points, $X_{\cap} \approx X$ and  $u \to 0$. We refer to the latter as a highly \emph{nonuniform corruption}. This measure is very general and can quantify various diverse corruptions. 

We can roughly divide the corruptions of ModelNet-C 
to \begin{enumerate}
    \item \emph{Uniform:} Scale, Rotation.
    \item \emph{Nonuniform:}  Drop-Global, Drop-Local, Add-Global, Add-Local.
\end{enumerate}
In the case of Jitter one can slightly extend the definition of uniformity by a ball of size $\epsilon$ around each point, when calculating $X_\cap$. Then uniformity grows with the standard deviation of the Jitter. 

\begin{figure}[!ptbh]
  \centering
   \includegraphics[width = 0.65
   \linewidth, height =0.48
   \linewidth]{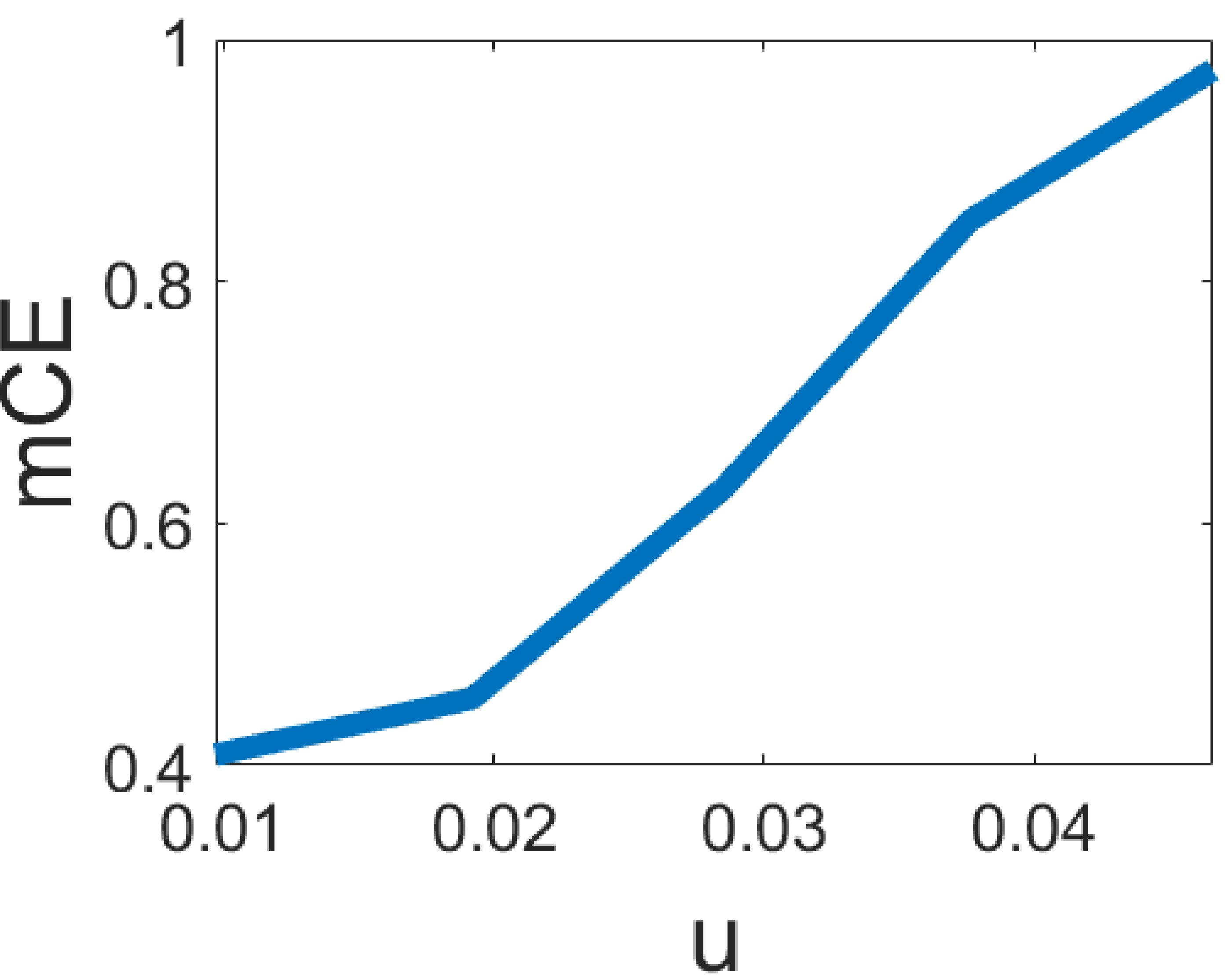}
   \caption{\textbf{mCE versus uniformity.} mCE is monotonically increasing with $u$ (average result on test set of ModelNet-C, Add-Global corruption, 5 degrees of severity).}
   \label{fig:add_global_plot}
\end{figure}

\begin{figure}[ptbh!]
  \centering
   \includegraphics[width = 
   \linewidth, height =0.68 
   \linewidth]{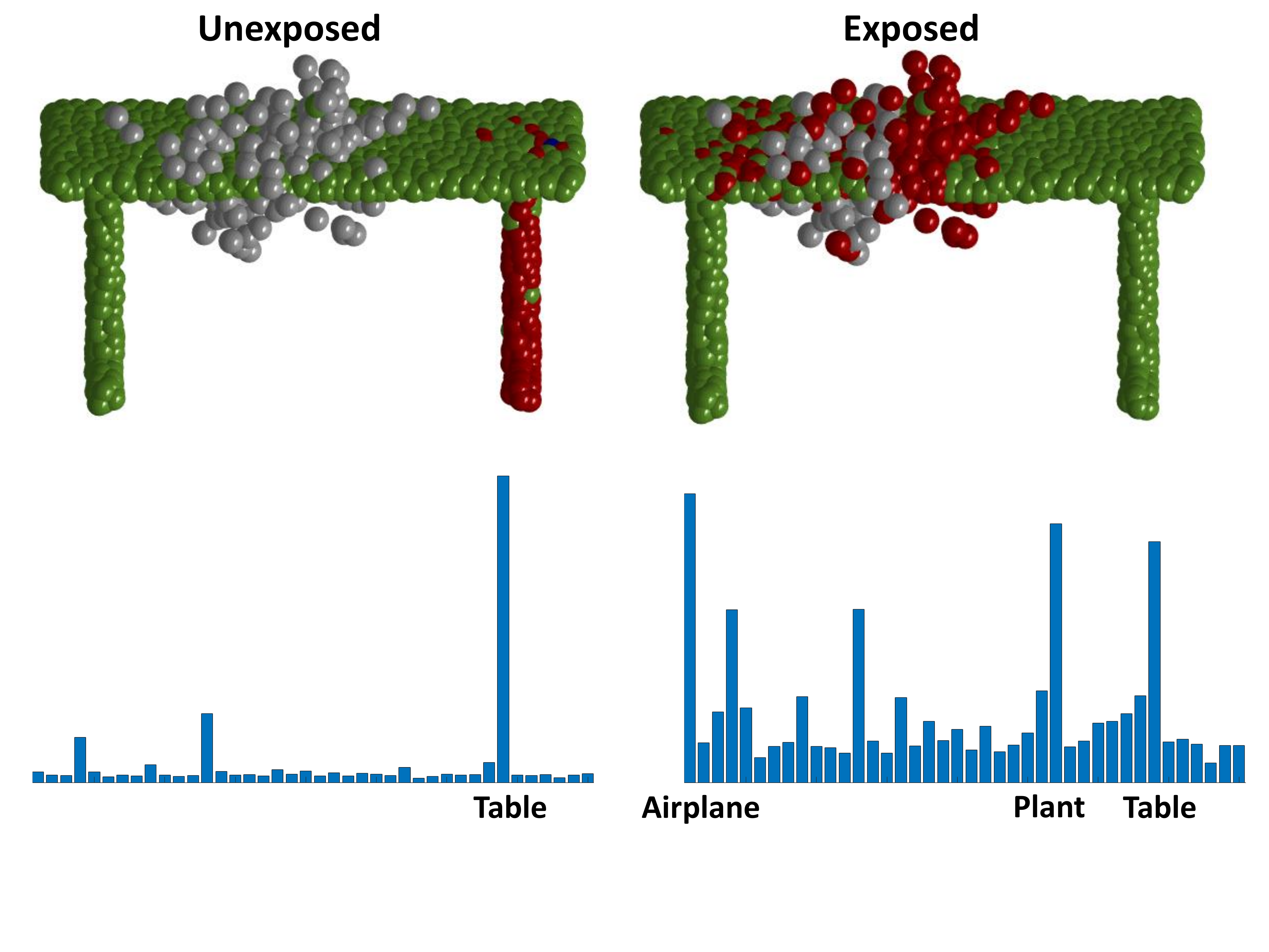}
   \caption{{\bf Exposure to corruption modeled as noise}. This illustration is based on the following experiment: All instances of class Table were corrupted by Add-Local. Curve is used for sampling. The sampled instances were divided into two groups: {\bf Unexposed} (left) are curves containing less than 10 corrupted points. {\bf Exposed} (right) are curves containing more than 50 corrupted points. Soft-Max predictions of each group were averaged (bottom row). The prediction of exposed curves is almost random (highly noisy) and can be well handled by averaging over the ensemble outputs. Unexposed instances are classified well.}
   \label{fig:noise}
\end{figure}

\emph{Our method performs best on nonuniform corruptions.}
In these cases some partial point clouds in the ensemble are mostly not exposed to the corruption, as shown in Fig. \ref{fig:corruptions}.
In these cases the classification is with high accuracy. Partial point clouds which are highly exposed to the corruption have high chances of yielding missclassification. However, our experiments indicate these missclassifications are quite random and can be approximately modeled as noise. As we perform a \emph{mean} operation over the ensemble outputs, this noise is mostly averaged out, where the correct (mostly unexposed) members of the ensemble dominate the decision. This phenomenon can be seen in the experiment shown and explained in Fig. \ref{fig:noise}. A plot of mCE as a function of $u$ is shown in Fig. \ref{fig:add_global_plot}, illustrating the above rationale.

\emph{The case of Jitter.}
Let us assume the points in the cloud are sampled approximately evenly, with a mean distance between each point of $\ell$. Let the Jitter corruption be of standard deviation $\sigma$.
If $\sigma \ll \ell$ essentially the point cloud is similar to the original clean one and most classifiers would perform well. In terms of the relaxed definition of uniformity, we can set $\epsilon = \ell$ and for $\sigma \ll \epsilon$ we get $u \to 0$ (the corruption is ''invisible'' to the classification network).
Random sampling copes very well with Jitter. We can view random sampling as reducing the point cloud resolution. Basically, we now have a larger distance $L > \ell$ between the points. Since the classifier is trained on the low resolution input, as long as $\sigma \ll L$ we get $u \to 0$. Hence, we are more robust to Jitter with a larger standard deviation. See \cite{pointguard} for additional perspectives and insights on this topic.

\subsection{Sampling, aggregation and network size} 
In Table \ref{table:aggregations} we show experimental results for each sub-sampling method with DGCNN\cite{dgcnn} as the basic model (additional models are shown in the Supp.). Ensembles of size four are tested, aggregated using mean. In addition, the results of all three methods (forming an ensemble of 12 members) are aggregated using either mean or majority-voting. 
Each sampling method has its strengths and weaknesses: Patches are much better in terms of accuracy on clean point clouds. In this case features are well preserved and patches have access to the full resolution within a region. 
Since patches have compact support, they perform better also on Drop-L and are somewhat immune to global corruptions (but less so, compared to the two other sampling methods). Curves perform best on Add-G. Intuitively, curves are most likely to sample regions with high conductance, thus global outliers are less likely to be selected. 
Random is excellent in the case of Jitter and generally has the best mCE among sampling methods.
The aggregated ensemble results, either using mean or majority-voting, are both better than any specific sampling method. Mean yields the best mCE, 
while majority voting, a popular ensemble aggregation method, turns out to be slightly worse (this trend is consistent when using other networks as well).

\begin{table}[!ptbh]
  \centering
    \begin{tabular}{p{2.8cm} || p{1.0cm} p{1.0cm} p{1.5cm}}
    \hline
    Network & OA ↑ & mCE ↓ & \#parameters\\
    \hline
    \bf{Classic DGCNN} & \bf{92.6\%} & \bf{1.000} & \bf{1.8M}\\
    \hline
    \bf{EPiC based on} &  &  & \\
    \hline
$\quad$ DGCNN-v1 & 88.7\% & 1.041 & 159K\\
 $\quad$   \bf{DGCNN-v2} & \bf{92.2\%} & \bf{0.773} & \bf{636K}\\ 
 $\quad$   DGCNN-v3 & 92.3\% & 0.720 & 2.47M\\ 
$\quad$    DGCNN-v4 & 93.0\% & 0.669 & 5.4M\\
$\quad$    DGCNN-v5 & 92.7\% & 0.684 & 39M\\  
    \hline
  \end{tabular}
  \caption{{\bf Network size vs Performance. 
  } EPiC with DGCNN-v2 is an excellent compromise of a lean architecture, with a small number of parameters, which is  much more robust (and almost as accurate), compared to the original full DGCNN network.}
  \label{table:shallowness}
\end{table}

{\bf Affect of network size.}
Since each ensemble member has access only to partial data , we check weather a full size model is required. Five  versions of DGCNN are examined, the most shallow is $v1$ and the deepest is $v5$, where $v4$ is the original network (architectures details appear in the Supp.). The results are shown in Table \ref{table:shallowness}. The number of parameters required for the entire ensemble (3 instances) is shown on the right column. Note that $v2$ yields a robust version with about third of the parameters of the classical network (top row), with just a slight degradation in overall accuracy.

\section{Conclusion}
\label{sec:conc}
In this work, we demonstrated the OOD robustness of ensembles based on partial information input for the task of point cloud classification. The approach relies on obtaining lowly-correlated input samples for each member of the ensemble. We integrate three types of sampling schemes: Curves, Patches and Random. In terms of training - only a single network for each sampling scheme is trained, which  saves training time. The ensemble is created in real time. This naturally increases inference time, but is reasonable for small ensembles (we show a size of $K=12$). For highly demanding real time applications, the networks can be duplicated and processed in parallel. For highly demanding memory consumption applications, significantly smaller networks can be used, which are still highly robust.
Note that ensembles can be distilled back to a single network, for better hardware and time efficiency, as suggested for instance in \cite{shen2019meal}.
Since our proposed approach is purely abstract it can be extended to additional problems. We plan to further investigate how such mechanisms can improve robustness in other fields as well.


{\small
\bibliographystyle{ieee_fullname}
\bibliography{egpaper_for_review}
}
\pagebreak
\section{Ablation Studies}
\begin{figure*}
 \centering
  
  \includegraphics[width=0.5\linewidth, height=0.2\linewidth]{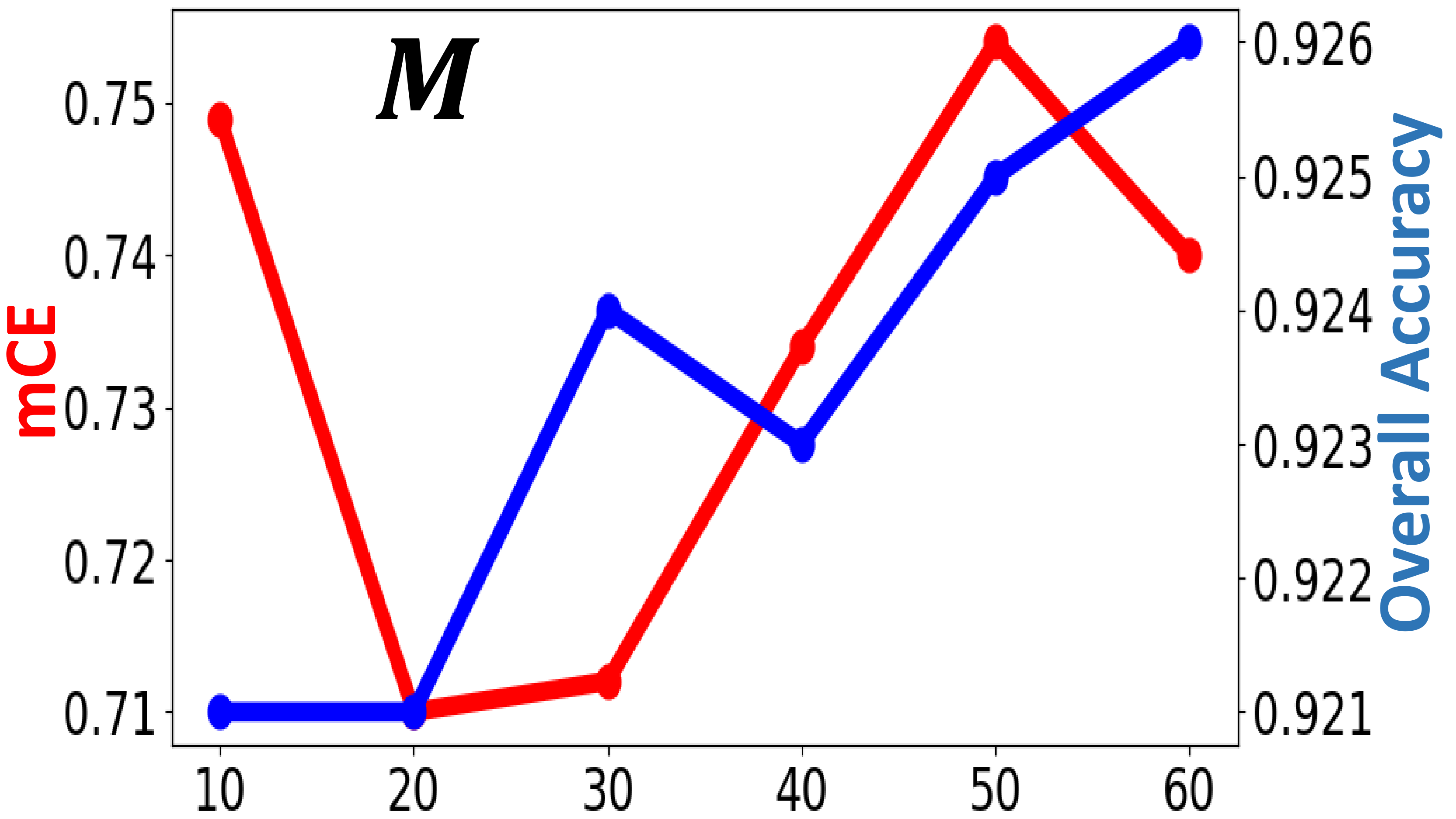}
  \caption{{\bf Neighbors in random walk, curve extraction}.}
  \label{fig:ablation_m}
\end{figure*}

\begin{figure*}
 \centering
  \includegraphics[width=0.5\linewidth, height=0.2\linewidth]{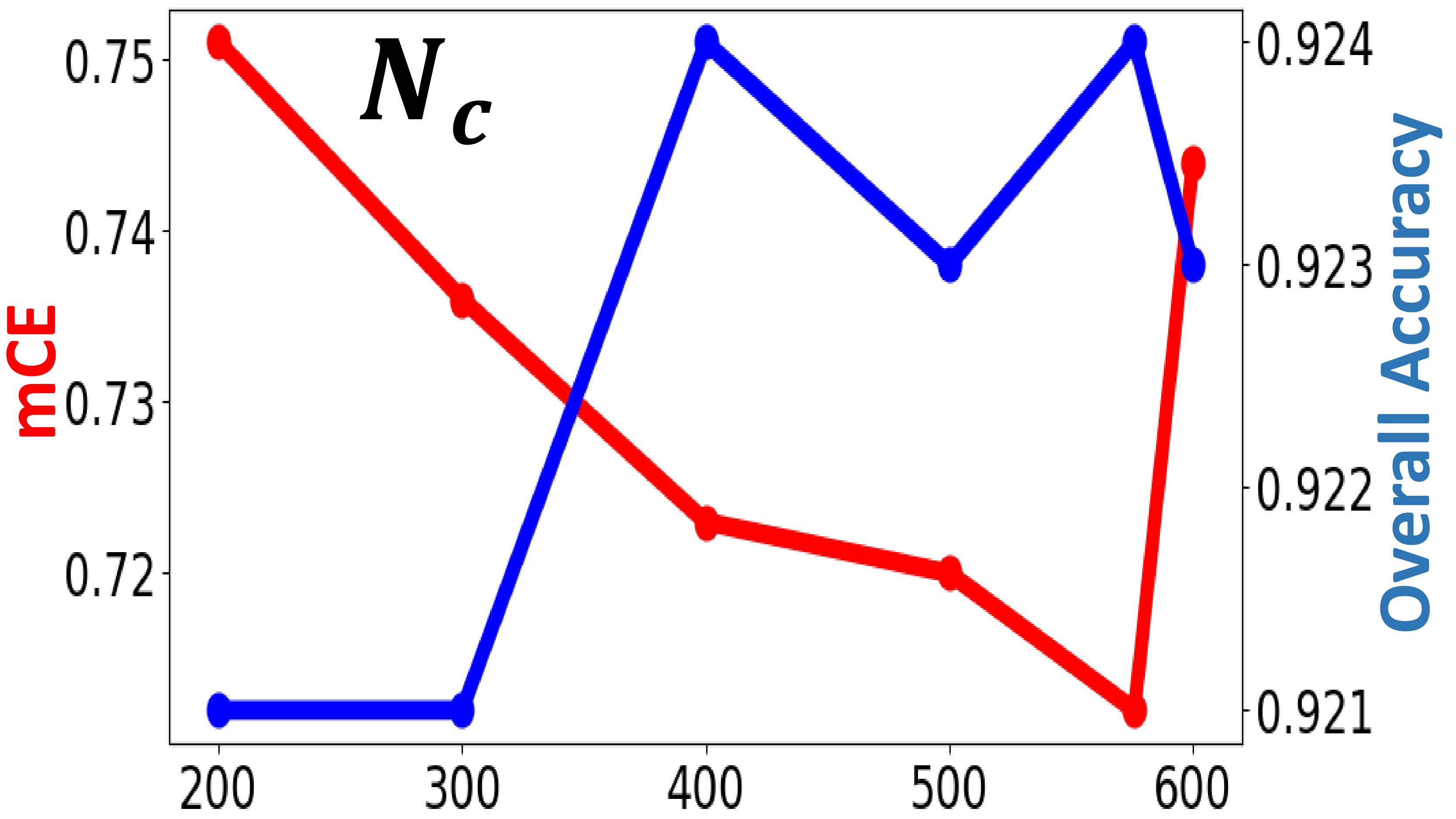}
  \caption{{\bf Curve sub-sample size}.}
  \label{fig:ablation_only_nc}
\end{figure*}

\begin{figure*}
 \centering
  \includegraphics[width=0.5\linewidth, height=0.2\linewidth]{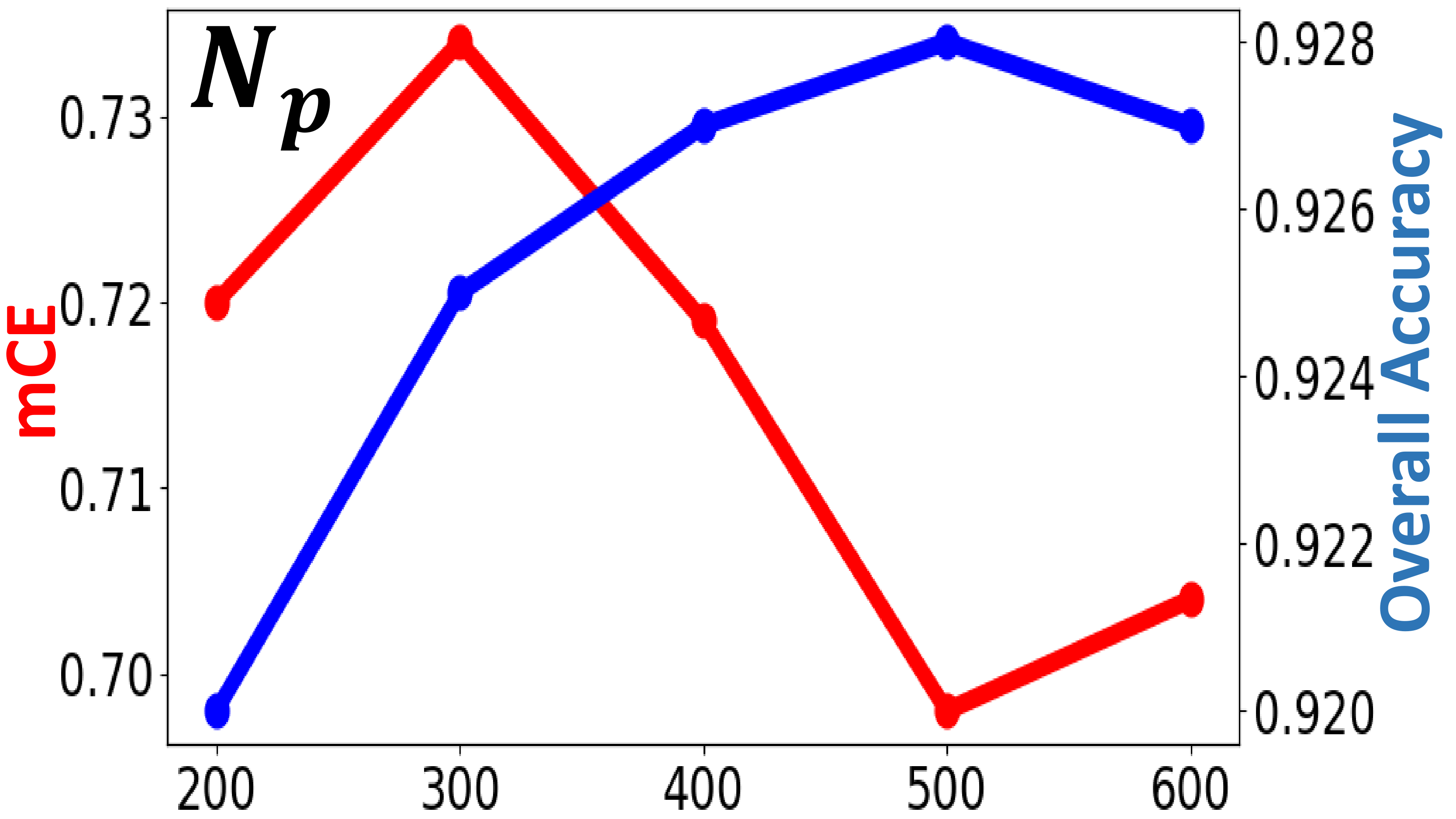}
  \caption{{\bf Patch sub-sample size}.}
  \label{fig:ablation_only_np}
\end{figure*}

\begin{figure*}
 \centering
  \includegraphics[width=0.5\linewidth, height=0.2\linewidth]{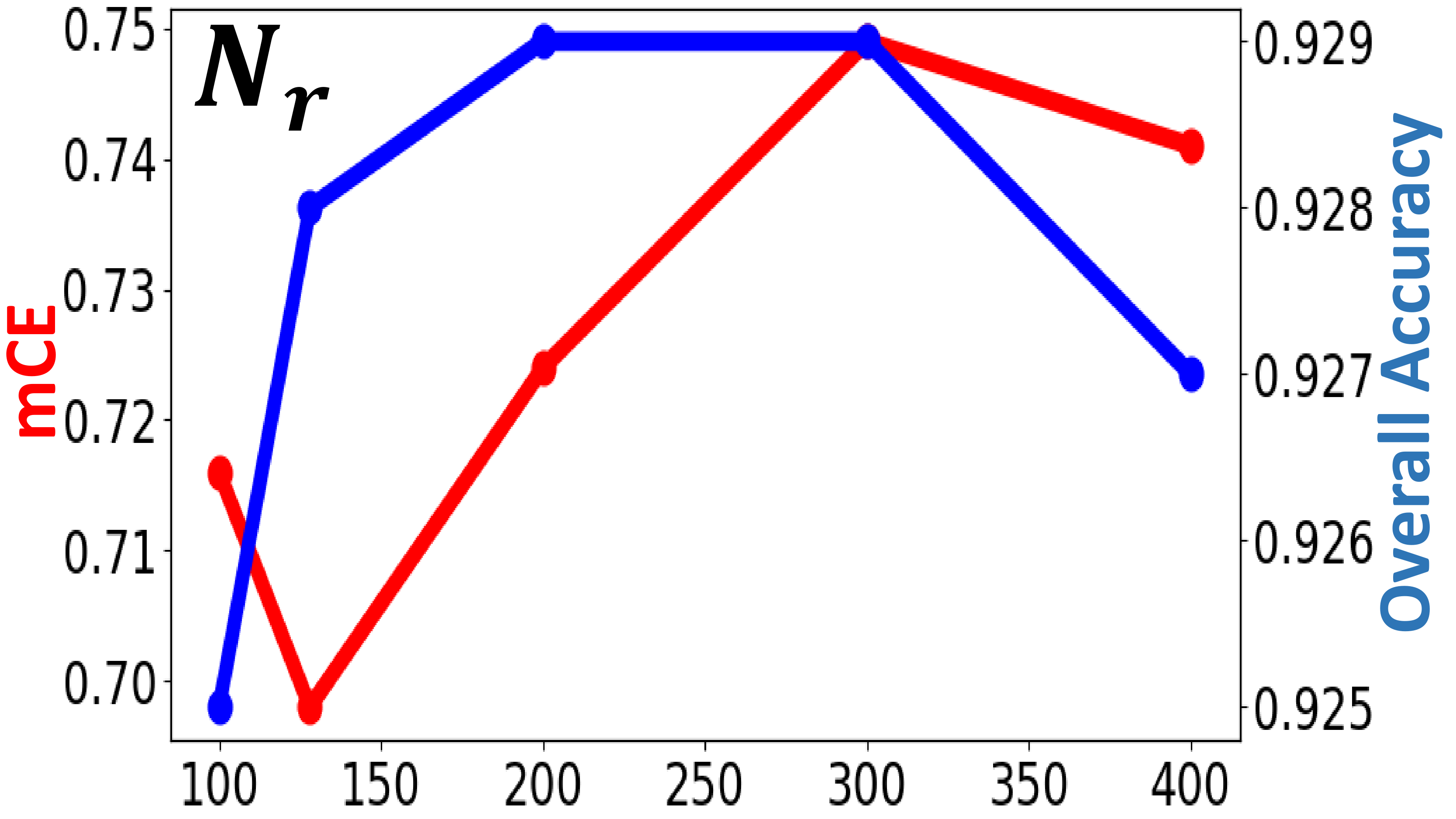}
  \caption{{\bf Random sub-sample size}.}
  \label{fig:ablation_only_nr}
\end{figure*}

\begin{figure*}
 \centering
  \includegraphics[width=0.5\linewidth, height=0.2\linewidth]{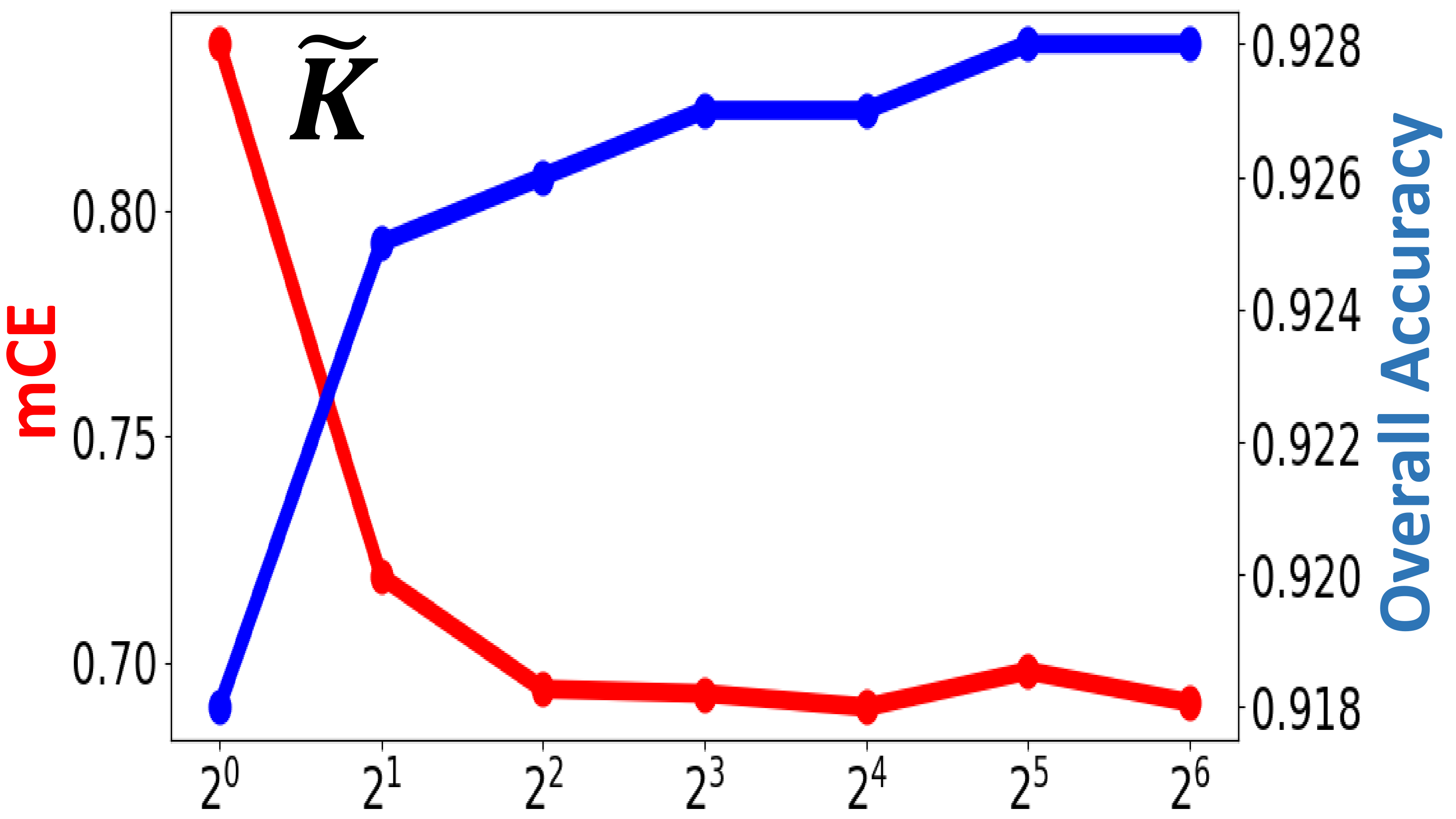}
  \caption{{\bf Ensemble size per sub-sample.
 }}
  \label{fig:ablation_only_k_tilde}
\end{figure*}

We conducted comprehensive ablative studies in order to better understand the impact of each hyper-parameter. The hyper-parameters are: Number of nearest neighbors for curve extraction, \textit{$M$}; Number of points in Curve sampling, \textit{$N_c$}; Number of points in Patch sampling, \textit{$N_p$}; Number of points in Random sampling, \textit{$N_r$} and number of partial point clouds for each sampling method, \textit{$\tilde{K}$} (we retained the restriction that $\tilde{K}$ is the same for all sampling methods).
Our parameter search was performed in the following manner: The base parameters were chosen to be $M=40$, $N_c=576$, $N_p=512$, $N_r=128$ and $\tilde{K}=32$. At each stage we  scanned the values of a specific parameter, while fixing all other parameters (the order is as listed above). 
We fixed the parameter to the optimal value after each stage. The optimized value was chosen manually by considering both overall accuracy and robustness to corruptions. Mean is used as an aggregating scheme.
The chosen values of hyper-parameters are: $M=40$, $N_c=512$, $N_p=512$, $N_r=128$ and $\tilde{K}=4$.
We observe that local sub-samples should be large enough to allow both accuracy and robustness. A global sub-sampling, such as random, requires far less points. Surprisingly, the ensemble size may be very small. Good results are obtained for $\tilde{K}=4$, allowing very reasonable inference time. 
See Figs \ref{fig:ablation_m}, \ref{fig:ablation_only_nc}, \ref{fig:ablation_only_np}, \ref{fig:ablation_only_nr} and \ref{fig:ablation_only_k_tilde}.

\begin{figure*}
  \centering
   \includegraphics[width=0.6\linewidth, height=0.7\linewidth]{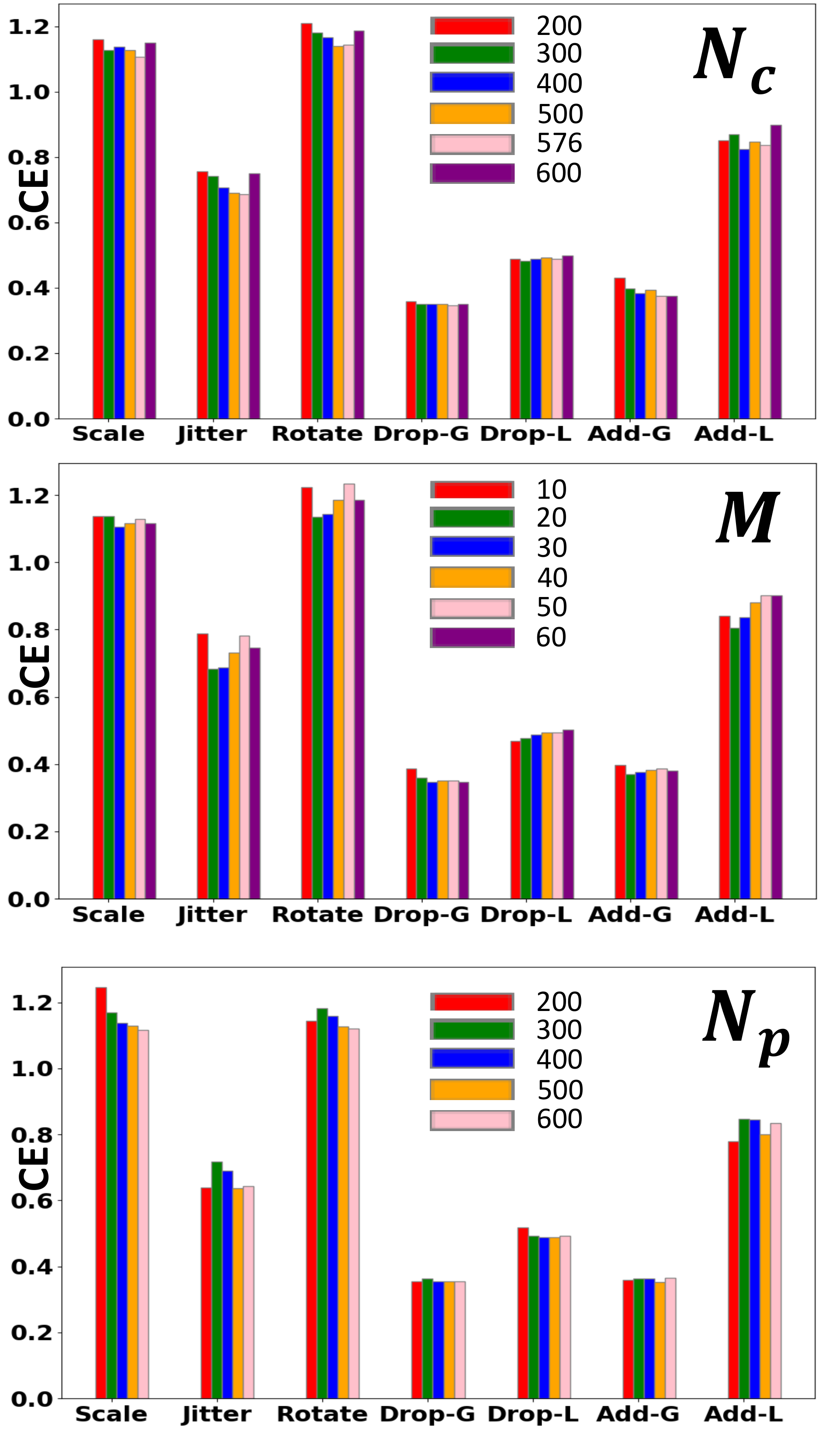}

   \caption{\bf{$N_c$, $M$ and $N_p$}} 
   \label{fig:ablation_nc_m_np}
\end{figure*}

\begin{figure*}
  \centering
   \includegraphics[width=0.6\linewidth, height=0.5\linewidth]{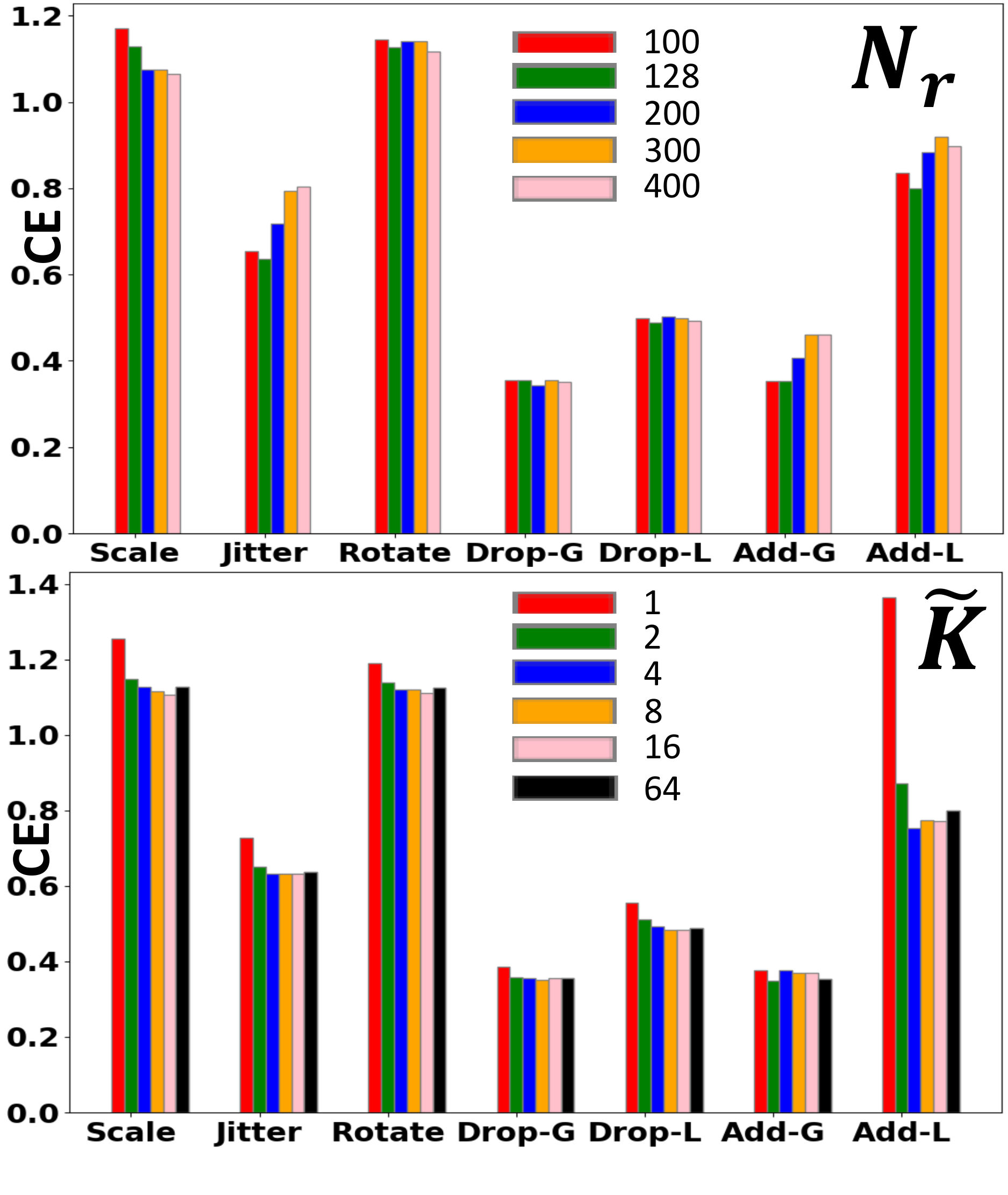}

   \caption{\bf{$N_r$ and $\tilde{K}$.}} 
   \label{fig:ablation_nr_k_tilde}
\end{figure*}

\textbf{Extended Ablation Studies}
We demonstrate the influence of each hyper-parameter under any of the corruptions in Figs \ref{fig:ablation_nc_m_np} and \ref{fig:ablation_nr_k_tilde}. The most meaningful hyper-parameter is $\tilde{K}$. Ensemble size of 12 yielding very good trade-off between running time, overall accuracy and robustness to OOD degradations. All corruptions are improved when enlarging the ensemble size, where Add-L is affected the most.





\section{Training procedure} We train all three partial point-cloud networks simultaneously and derivate each sub-sampling network's loss with regard to the entire point-cloud label as detailed in Algorithm \ref{alg:training}

\section{Sub-Samples analysis of more models}
\textbf{GDANet.} For training, We lower the batch size to 50 and changed the GDM Module to accept point clouds with less than 256 elements by lower M value to 128 in Geometry-Disentangle Module (GDM).

All other networks did not require any extra tweaks, the analysis results of all networks is detailed in Tables
\ref{table:gdanet_unaugmented_sub_samples},
\ref{table:gdanet_augmented_sub_samples},
\ref{table:dgcnn_unaugmented_sub_samples}, 
\ref{table:dgcnn_augmented_sub_samples}, 
\ref{table:pct_unaugmented_sub_samples},
\ref{table:pct_augmented_sub_samples}, 
\ref{table:rpc_unaugmented_sub_samples} and
\ref{table:rpc_augmented_sub_samples}. 

\section{Network size (shallowness) analysis}

DGCNN is constructed from 4 levels of depth. Features of current level are recalculated based on features of previous level. In order to examine the required depth for our case we used different versions of DGCNN. Each version is constructed using a different number of graph reconstructions. For example, DGCNN-v1 is built only from one graph based on the spatial distance, therefore features are directly derived from  spatial information. As depth increases (DGCNN-v2 through DGCNN-v5) the network acquires additional, more complex,  semantic contextual information. DGCNN-v1 has up to 64 embedding dimensions, DGCNN-v2 up to 128, and so on. Each version doubles the embedding dimensions  (which grows exponentially with depth).
Comparing to the classical DGCNN (v4), we observe  that EPiC with DGCNN-v2 yields a highly robust network (MCE=0.773 compare to 1.000) with much less learnable parameters (636K compare to 1.8M). This comes at the cost of a slight accuracy drop (92.2\% compared to 92.6\%). Thus, the EPiC framework can be very lean and economic from a production perspective.

\section{Corruptions and sub-sampling}

\begin{figure*}[h]
  \centering
   \includegraphics[width=1\linewidth, height=0.75\linewidth]{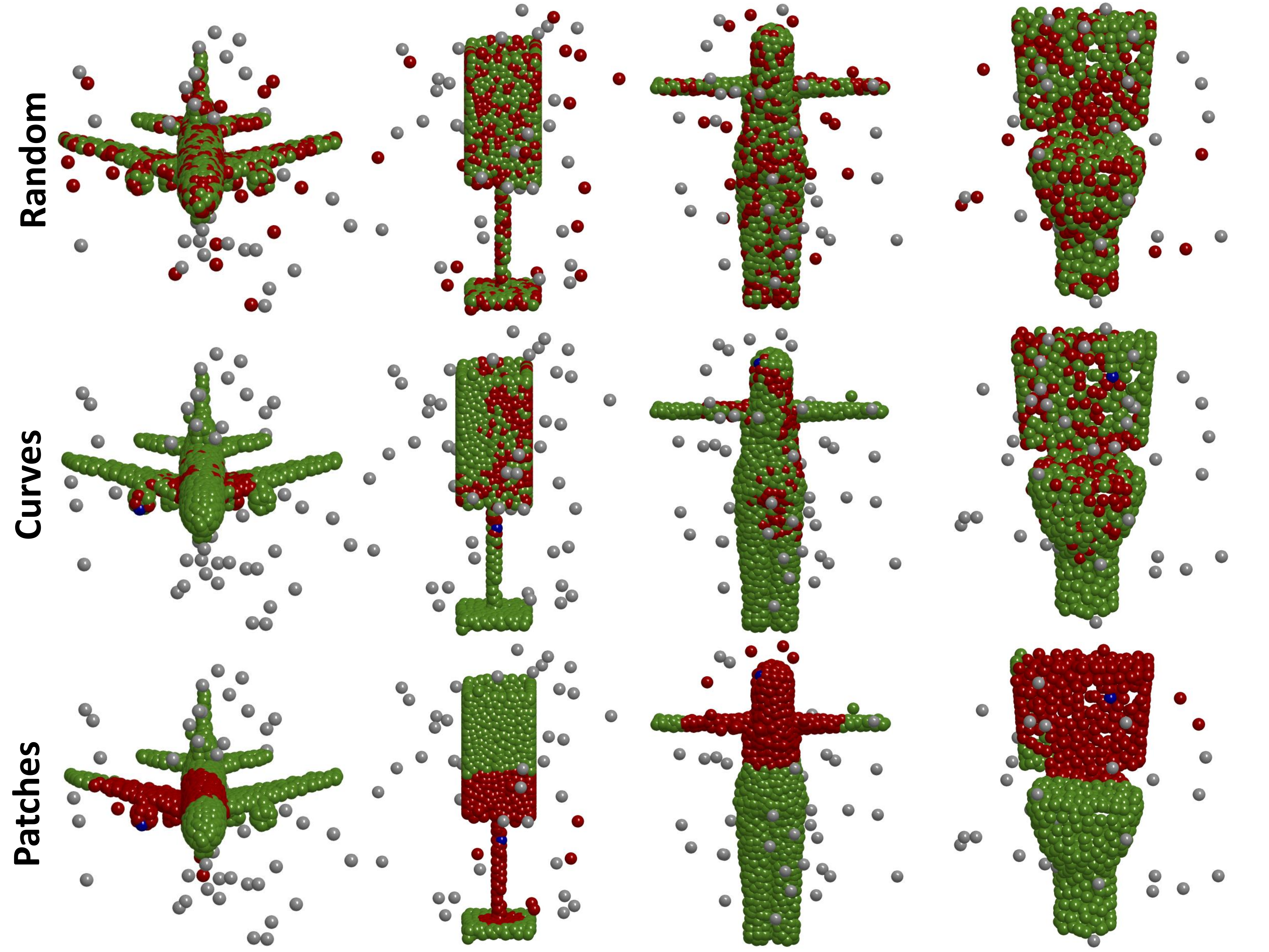}

   \caption{\bf{Add Global}} 
   \label{fig:supp_add_g}
\end{figure*}

\begin{figure*}[h]
  \centering
   \includegraphics[width=1\linewidth, height=0.75\linewidth]{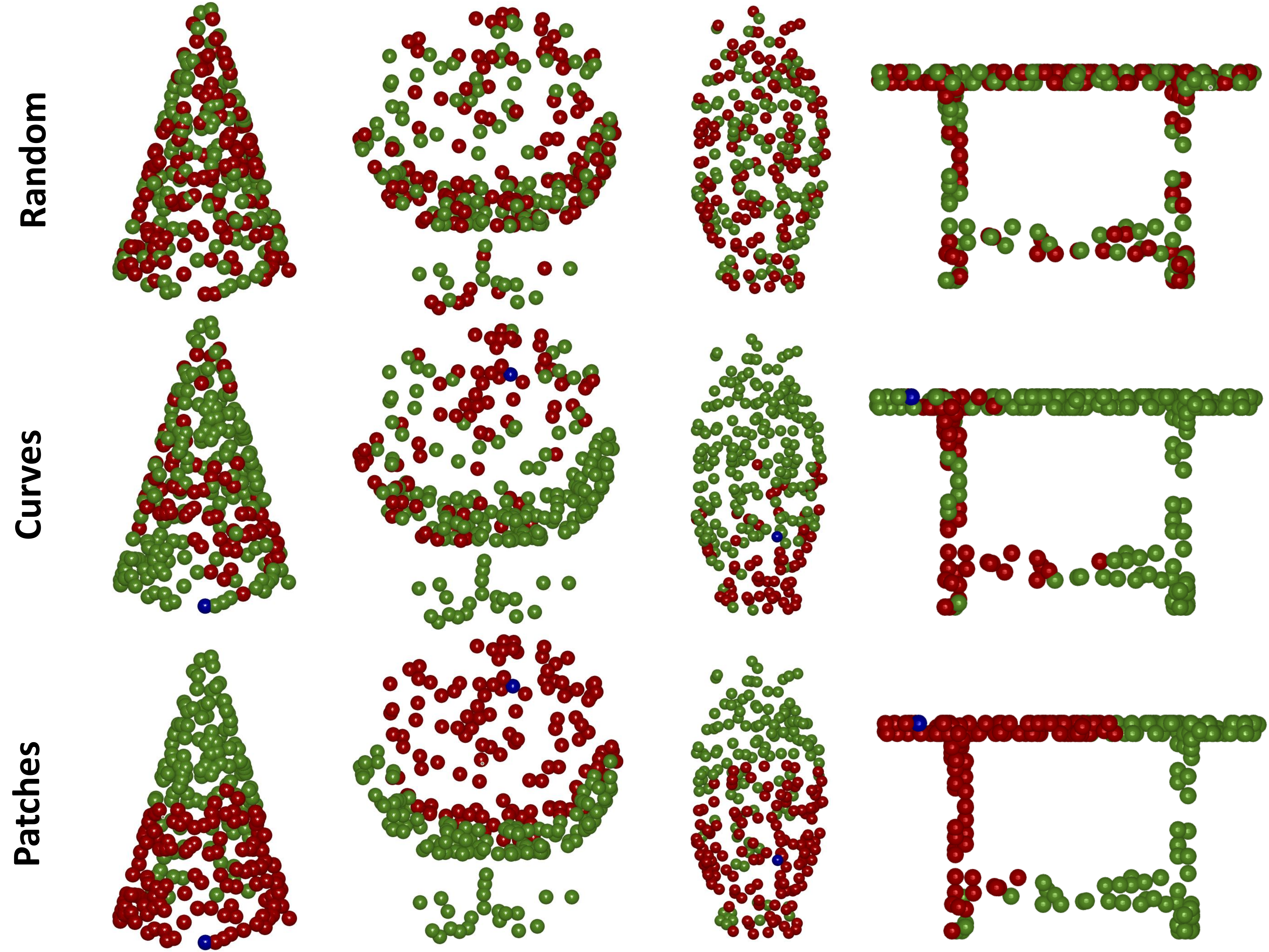}

   \caption{\bf{Drop Global}} 
   \label{fig:supp_drop_g}
\end{figure*}

\begin{figure*}[h]
  \centering
   \includegraphics[width=1\linewidth, height=0.75\linewidth]{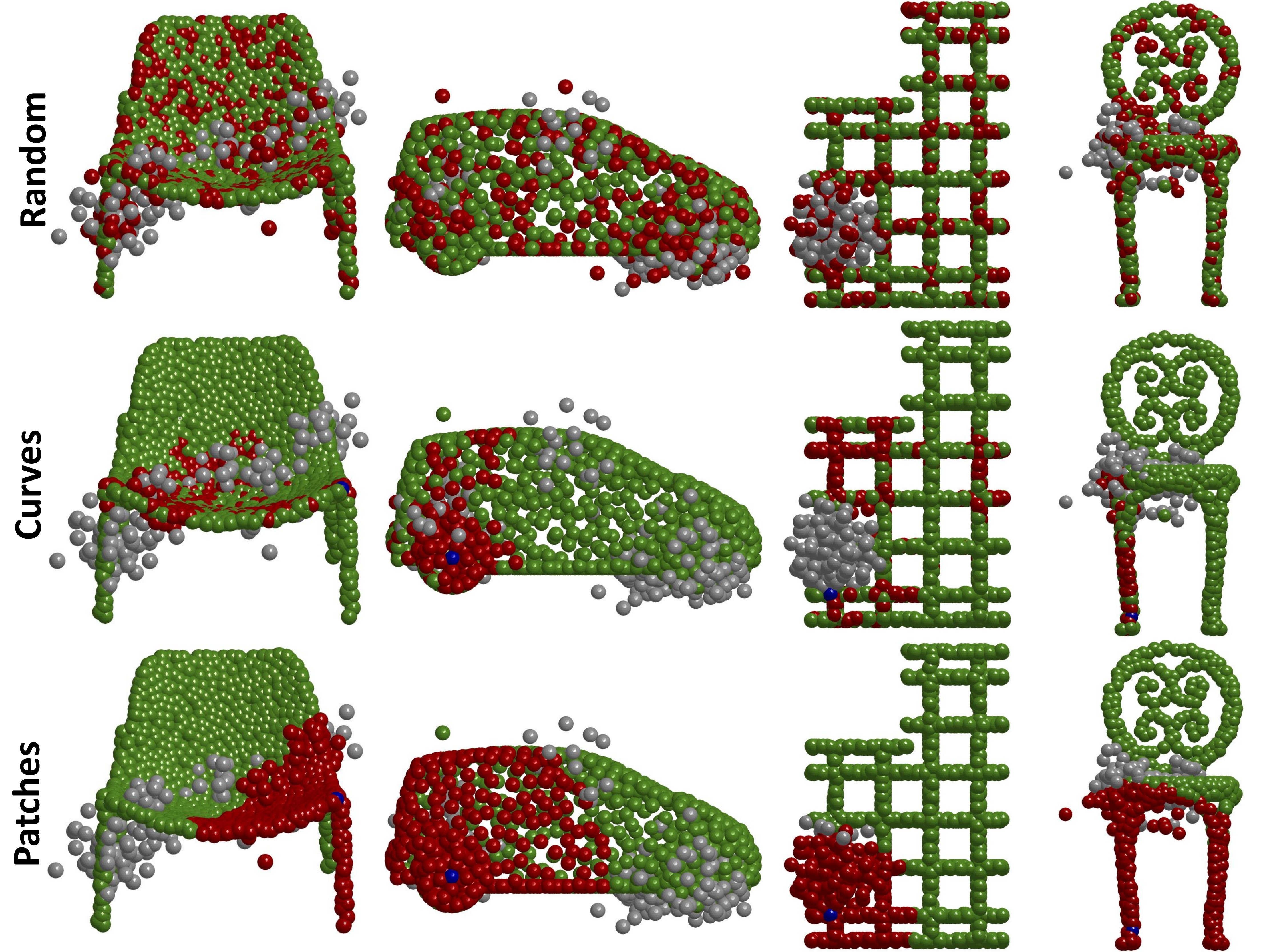}

   \caption{{\bf Add Local.}
   }
   \label{fig:supp_add_l}
\end{figure*}

\begin{figure*}[h]
  \centering
   \includegraphics[width=1\linewidth, height=0.75\linewidth]{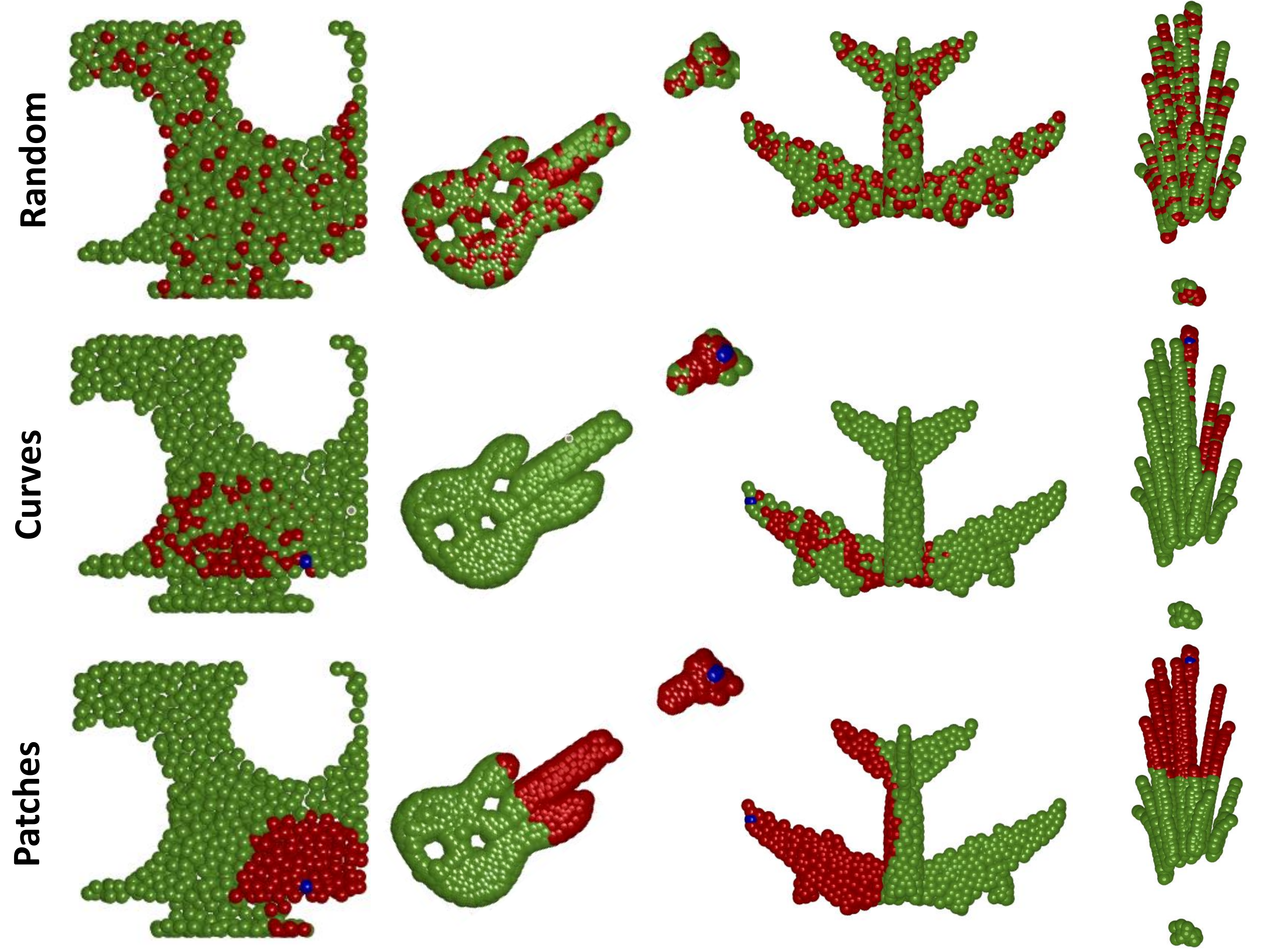}

   \caption{{\bf Drop Local.}
   }
   \label{fig:supp_drop_l}
\end{figure*}

Here we give some of our insights on how the different sub-sampling methods react to different corruptions. Examples are shown in Figs. \ref{fig:supp_add_g}, \ref{fig:supp_drop_g}, \ref{fig:supp_add_l} and \ref{fig:supp_drop_l}.

{\bf Add global.} Curves are highly insensitive to relatively far points (which often appear in Add-G). Therefore, they perform best under this corruption. Patches and Random are not influenced by the conductance, thus the added points may be considered as well, corrupting the sample. 

{\bf Add local.} Here, the corruption is smoother in space and may be interpreted as part of the shape. Patches and Curves are more likely to be affected by it. 

{\bf Drop Global.} Drop Global can be viewed as globally changing the density of the points of the shape. Therefore, it makes sense that there is a direct proportion between sub-sampling performance and typical density (Patches are dense while Random is sparse, Curves are somewhere in between).
Moreover, we take into account a variable number of points, e.g. for Patches $min(N,N_p)$. Thus, when the number of points is lower, as in the case of Drop-Global, Patches cover more of the shape.

{\bf Drop Local.} This corruption may cause separation between different shape parts (creating disconnected regions). Curves may thus be ``stuck'' in a disconnected region. Patches are less affected by this corruption and perform better.

\subsection{Networks and augmentation evaluation}
Evaluation of most augmented and un-augmented networks was reproduced by the code supplied by Ren et al. Additionally, in the ensemble section we show results on PointGuard. These experiments are detailed below. 


\textbf{Point Guard.} This method obtains highly sparse random sub-samples (they suggest $N=16$), using a large ensemble of size $K=1,000$. Its theoretical analysis and experimental setting do not assume a specific architecture for the classifier. Thus,  DGCNN is used as the basic classifier. As suggested in PointGuard we trained our classifier for 7500 epochs to work on the sub-samples, obtaining an overall accuracy of $80.5\%$ (estimated on randomly 4 sub-samples per each sample in the test set). Predictions are aggregated using majority voting. 
Note that to obtain provable robustness they use very small $N$ and try to compensate by extremely large $K$. However this tradeoff turns out to be significantly inferior in terms of overall accuracy and less competitive in $mCE$.

\textbf{WolfMix.} We follow ModelNet-C evaluation metrics.
we use the default hyper-parameters in PointWOLF (Kim et al., 2021). We set the number of
anchors to 4, sampling method to farthest point sampling, kernel bandwidth to 0.5, maximum local rotation range to 10 degrees, maximum local scaling to 3, and maximum local translation to 0.25. AugTune proposed along with PointWOLF is not used in training. For the mixing step, we use the default hyper-parameters in RSMix (Lee et al., 2021). We set RSMix probability to 0.5, $\beta$ to 1.0, and the maximum number of point modifications to 512. For training, the number of neighbors in k-NN is reduced to 20, the number of epochs is increased to 500.

\section{On Batch Normalization and OOD Robustness}
Using fixed learnt batch normalization parameters at test time can be problematic, since it assumes that the samples of the train and test are both approximately of similar  distributions. However, different types of corruptions can yield different statistics.
Obviously, learning the statistics at test time may violate the OOD principle. Nevertheless, for some ``offline'' applications, which can work in batches, we wanted to examine whether test time batch normalization is helpful  
and increases robustness. To demonstrate this idea, normalization parameters were computed during evaluation (with a batch size of 256). We observe that $mCE$ significantly improves, as can be seen in Table \ref{table:batchnorm}.

    

\begin{table*}
  \centering
    \begin{tabular}{p{4cm} || p{1.1cm} p{1.1cm} p{1.1cm} p{1.1cm} p{1.2cm} p{1.2cm} p{1.1cm} p{1.1cm} p{1.1cm}}
    \hline
    Sub-samples & OA ↑ & mCE ↓ & Scale & Jitter & Drop-G & Drop-L & Add-G & Add-L & Rotate \\
    \hline
    GDANet-Curves (\#4) & 91.3\% & 1.047 & 1.298 & 1.551 & 0.423 & 0.599 & 0.336 & 1.549 & 1.572\\
    GDANet-Patches (\#4) & \underline{93.2\%} & 0.861 & \underline{0.957} & 1.358 & 0.504 & 0.551 & 0.481 & 1.044 & \textbf{1.130}\\
    GDANet-Random (\#4) & 90.9\% & 0.819 & 1.287 & \textbf{0.462} & 0.359 & 0.836 & 0.512 & 0.898 & 1.381\\
    \hline
    GDANet-Mean (\#12) & \textbf{93.6\%} & \textbf{0.704} & \textbf{0.936} & \underline{0.864} & \textbf{0.315} &  \textbf{0.478} & \textbf{0.295} & \textbf{0.862} &  \underline{1.177}\\
    GDANet-Maj. Vot.(\#12) & \underline{93.2\%} & \underline{0.749} & 0.968 & 1.028 & \underline{0.343} &  \underline{0.498} & \underline{0.325} & \underline{0.876} &  1.205\\
    
    \hline
  \end{tabular}
  \caption{{\bf Un-Augmented GDANet sub-samples analysis.} {\bf Bold} best. \underline{Underline} second best.}
  \label{table:gdanet_unaugmented_sub_samples}
\end{table*}

\begin{table*}
  \centering
    \begin{tabular}{p{4cm} || p{1.1cm} p{1.1cm} p{1.1cm} p{1.1cm} p{1.2cm} p{1.2cm} p{1.1cm} p{1.1cm} p{1.1cm}}
    \hline
    Sub-samples & OA ↑ & mCE ↓ & Scale & Jitter & Drop-G & Drop-L & Add-G & Add-L & Rotate \\
    \hline
    GDANet-Curves (\#4) & 90.8\% & 0.740 & 1.266 & 0.984 & 0.411 & 0.614 & 0.332 & 0.818 & 0.758\\
    GDANet-Patches (\#4) & 92.1\% & 0.667 & \underline{0.979} & 1.275 & 0.464 & 0.493 & 0.353 & 0.531 & \textbf{0.572}\\
    GDANet-Random (\#4) & 90.8\% & 0.646 & 1.234 & \textbf{0.462} & 0.383 & 0.758 & 0.359 & 0.571 & 0.758\\
    \hline
    GDANet-Mean (\#12) & \textbf{92.5\%} & \textbf{0.530} & \textbf{0.968} & \underline{0.639} & \textbf{0.343} &  \textbf{0.473} & \textbf{0.275} & \textbf{0.433} &  \underline{0.577}\\
    GDANet-Maj. Vot.(\#12) & \underline{92.2}\% & \underline{0.558} & 1.000 & 0.725 & \underline{0.355} &  \underline{0.478} & \underline{0.292} & \underline{0.462} &  0.591\\
    
    \hline
  \end{tabular}
  \caption{{\bf WolfMix Augmented GDANet sub-samples analysis.} {\bf Bold} best. \underline{Underline} second best.}
  \label{table:gdanet_augmented_sub_samples}
\end{table*}

\begin{table*}
  \centering
    \begin{tabular}{p{4cm} || p{1.1cm} p{1.1cm} p{1.1cm} p{1.1cm} p{1.2cm} p{1.2cm} p{1.1cm} p{1.1cm} p{1.1cm}}
    \hline
    Sub-samples & OA ↑ & mCE ↓ & Scale & Jitter & Drop-G & Drop-L & Add-G & Add-L & Rotate \\
    \hline
    DGCNN-Curves (\#4) & 90.7\% & 1.069 & 1.628 & 1.297 & 0.431 & 0.729 & \underline{0.363} & 1.618 & 1.414\\
    DGCNN-Patches (\#4) & \underline{92.8\%} & 0.793 & \textbf{0.989} & 1.165 & 0.577 & 0.536 & 0.505 & 0.851 & \textbf{0.930}\\
    DGCNN-Random (\#4) & 91.5\% & 0.766 & 1.234 & \textbf{0.399} & \underline{0.351} & 0.812 & 0.580 & \textbf{0.793} & 1.195\\
    \hline
    DGCNN-Mean(\#12) & \textbf{93.0\%} & \textbf{0.669} & \underline{1.000} &  \underline{0.680} & \textbf{0.331} &  \textbf{0.498} & \textbf{0.349} & 0.807 &  \underline{1.019}\\
    DGCNN-Maj. Vot.(\#12) & 92.6\% & \underline{0.706} & 1.043 &  0.794 & 0.359 &  \underline{0.517} & 0.380 & \underline{0.800} &  1.051\\
    
    \hline
  \end{tabular}
  \caption{{\bf Un-Augmented DGCNN sub-samples analysis.} {\bf Bold} best. \underline{Underline} second best.}
  \label{table:dgcnn_unaugmented_sub_samples}
\end{table*}

\begin{table*}
  \centering
    \begin{tabular}{p{4cm} || p{1.1cm} p{1.1cm} p{1.1cm} p{1.1cm} p{1.2cm} p{1.2cm} p{1.1cm} p{1.1cm} p{1.1cm}}
    \hline
    Sub-samples & OA ↑ & mCE ↓ & Scale & Jitter & Drop-G & Drop-L & Add-G & Add-L & Rotate \\
    \hline
    DGCNN-Curves (\#4) & 89.5\% & 0.802 & 1.426 & 0.896 & 0.468 & 0.676 & 0.386 & 0.873 & 0.888\\
    DGCNN-Patches (\#4) & 91.7\% & 0.618 & \underline{1.053} & 0.823 & 0.536 & 0.512 & 0.380 & 0.433 & \textbf{0.591}\\
    DGCNN-Random (\#4) & 91.2\% & 0.639 & 1.298 & \textbf{0.405} & \underline{0.379} & 0.768 & 0.369 & 0.527 & 0.730\\
    \hline
    DGCNN-Mean (\#12) & \underline{92.1\%} & \textbf{0.529} & \textbf{1.021} &  \underline{0.541} & \textbf{0.355} &  \textbf{0.488} & \textbf{0.288} & \textbf{0.407} &  \underline{0.600}\\
    DGCNN-Maj. Vot.(\#12) & \textbf{92.3\%} & \underline{0.552} & \underline{1.053} &  0.592 & \underline{0.379} &  \underline{0.498} & \underline{0.308} & \underline{0.418} &  0.614\\
    
    \hline
  \end{tabular}
  \caption{{\bf WolfMix Augmented DGCNN sub-samples analysis.} {\bf Bold} best. \underline{Underline} second best.}
  \label{table:dgcnn_augmented_sub_samples}
\end{table*}

\begin{table*}
  \centering
    \begin{tabular}{p{4cm} || p{1.1cm} p{1.1cm} p{1.1cm} p{1.1cm} p{1.2cm} p{1.2cm} p{1.1cm} p{1.1cm} p{1.1cm}}
    \hline
    Sub-samples & OA ↑ & mCE ↓ & Scale & Jitter & Drop-G & Drop-L & Add-G & Add-L & Rotate \\
    \hline
    PCT-Curves (\#4) & 91.1\% & 0.934 & 1.234 & 1.320 & 0.391 & 0.551 & 0.322 & 1.393 & 1.330\\
    PCT-Patches (\#4) & 92.5\% & 0.915 & 0.989 & 1.557 & 0.556 & 0.541 & 0.549 & 1.080 & 1.135\\
    PCT-Random (\#4) & 91.9\% & 0.741 & 1.245 & \textbf{0.427} & 0.335 & 0.744 & 0.502 & 0.785 & 1.149\\
    \hline
    PCT-Mean (\#12) & \textbf{93.4\%} & \textbf{0.646} & \textbf{0.894} & \underline{0.851} & \textbf{0.306} &  \textbf{0.435} & \textbf{0.285} & \textbf{0.735} &  \textbf{1.019}\\
    PCT-Maj. Vot.(\#12) & \underline{93.1\%} & \underline{0.693} & \underline{0.957} & 0.994 & \underline{0.331} &  \underline{0.454} & \underline{0.315} & \underline{0.760} &  \underline{1.042}\\
    
    \hline
  \end{tabular}
  \caption{{\bf Un-Augmented PCT sub-samples analysis.} {\bf Bold} best. \underline{Underline} second best.}
  \label{table:pct_unaugmented_sub_samples}
\end{table*}

\begin{table*}
  \centering
    \begin{tabular}{p{4cm} || p{1.1cm} p{1.1cm} p{1.1cm} p{1.1cm} p{1.2cm} p{1.2cm} p{1.1cm} p{1.1cm} p{1.1cm}}
    \hline
    Sub-samples & OA ↑ & mCE ↓ & Scale & Jitter & Drop-G & Drop-L & Add-G & Add-L & Rotate \\
    \hline
    PCT-Curves (\#4) & 90.4\% & 0.699 & 1.255 & 0.927 & 0.427 & 0.570 & 0.332 & 0.698 & 0.684\\
    PCT-Patches (\#4) & \textbf{92.7\%} & 0.633 & \textbf{0.904} & 1.339 & 0.391 & \textbf{0.420} & 0.312 & 0.505 & \underline{0.558}\\
    PCT-Random (\#4) & 91.0\% & 0.636 & 1.245 & \textbf{0.554} & 0.375 & 0.700 & 0.342 & 0.567 & 0.670\\
    \hline
    PCT-Mean (\#12) & \textbf{92.7\%} & \textbf{0.510} & \underline{0.915} & \underline{0.699} & \textbf{0.323} &  \underline{0.425} & \textbf{0.268} & \textbf{0.404} &  \textbf{0.535}\\
    PCT-Maj. Vot.(\#12) & \underline{92.6\%} & \underline{0.532} & 0.947 & 0.756 & \underline{0.343} &  0.430 & \underline{0.271} & \underline{0.422} &  \underline{0.558}\\
    
    \hline
  \end{tabular}
  \caption{{\bf WolfMix Augmented PCT sub-samples analysis.} {\bf Bold} best. \underline{Underline} second best.}
  \label{table:pct_augmented_sub_samples}
\end{table*}

\begin{table*}
  \centering
    \begin{tabular}{p{4cm} || p{1.1cm} p{1.1cm} p{1.1cm} p{1.1cm} p{1.2cm} p{1.2cm} p{1.1cm} p{1.1cm} p{1.1cm}}
    \hline
    Sub-samples & OA ↑ & mCE ↓ & Scale & Jitter & Drop-G & Drop-L & Add-G & Add-L & Rotate \\
    \hline
    RPC-Curves (\#4) & 91.6\% & 1.068 & 1.287 & 1.680 & 0.399 & 0.594& 0.322 & 1.495 & 1.702\\
    RPC-Patches (\#4) & 92.4\% & 0.934 & 0.979 & 1.532 & 0.488 & 0.536 & 0.498 & 1.233 & \textbf{1.274}\\
    RPC-Random (\#4) & 91.5\% & 0.804 & 1.181 & \textbf{0.491} & 0.355 & 0.739 & 0.498 & \textbf{0.855} & 1.512\\
    \hline
    RPC-Mean (\#12) & \textbf{93.6\%} & \textbf{0.750} & \textbf{0.915} & \underline{1.057} & \textbf{0.323} &  \textbf{0.440} & \textbf{0.281} & \underline{0.902} &  \underline{1.330}\\
    RPC-Maj. Vot.(\#12) & \underline{93.0\%} & \underline{0.791} & \underline{0.957} & 1.168 & \underline{0.343} &  \underline{0.473} & \underline{0.319} & 0.913 &  1.367\\
    
    \hline
  \end{tabular}
  \caption{{\bf Un-Augmented RPC sub-samples analysis.} {\bf Bold} best. \underline{Underline} second best.}
  \label{table:rpc_unaugmented_sub_samples}
\end{table*}

\begin{table*}
  \centering
    \begin{tabular}{p{4cm} || p{1.1cm} p{1.1cm} p{1.1cm} p{1.1cm} p{1.2cm} p{1.2cm} p{1.1cm} p{1.1cm} p{1.1cm}}
    \hline
    Sub-samples & OA ↑ & mCE ↓ & Scale & Jitter & Drop-G & Drop-L & Add-G & Add-L & Rotate \\
    \hline
    RPC-Curves (\#4) & 91.7\% & 0.686 & 1.106 & 1.038 & 0.375 & 0.551 & 0.315 & 0.698 & 0.716\\
    RPC-Patches (\#4) & \underline{92.2\%} & 0.603 & 0.957 & 1.190 & 0.399 & \underline{0.430} & 0.298 & 0.415 & \textbf{0.535}\\
    RPC-Random (\#4) & 91.2\% & 0.609 & 1.170 & \textbf{0.459} & 0.359 & 0.662 & 0.342 & 0.542 & 0.730\\
    \hline
    RPC-Mean (\#12) & \textbf{92.7\%} & \textbf{0.501} & \textbf{0.915} & \underline{0.680} & \textbf{0.315} &  \textbf{0.420} & \textbf{0.251} & \textbf{0.382} &  \underline{0.544}\\
    RPC-Maj. Vot.(\#12) &\textbf{92.7\%} & \underline{0.526} & \underline{0.947} & 0.766 & \underline{0.335} &  \textbf{0.420} & \underline{0.268} & \underline{0.400} &  0.549\\
    
    \hline
  \end{tabular}
  \caption{{\bf WolfMix Augmented RPC sub-samples analysis.} {\bf Bold} best. \underline{Underline} second best.}
  \label{table:rpc_augmented_sub_samples}
\end{table*}

\algrenewcomment[1]{\(\triangleright\) #1}
\algnewcommand{\LineComment}[1]{\State \(\triangleright\) #1}

\begin{algorithm*}[h]
\caption{Classification using EPiC (Training)}\label{alg:training}
\begin{algorithmic}

\For{$epoch \in epochs$} 
\For{$X, label \in TrainingDataSet$} 

\State $anchors \gets RandomlySelect([0:1023])$

\For{$anchor \in anchors$} 
\LineComment{Fetch partial point clouds}
\State $Patch \gets FetchPatch(X, anchors(k))$
\State $Curve \gets FetchCurve(X, anchors(k))$
\State $Random \gets FetchRandom(X)$

\LineComment{Apply models}
\State $P_{Patch}\gets model_{Patches}(Patch)$
\State $P_{Curve} \gets model_{Curves}(Curve)$
\State $P_{Random} \gets model_{Random}(Random)$

\LineComment{Derivate with regard to the entire point-cloud label}
\State $Loss_{Patch} \gets backward(P_{Patch}, label)$ 
\State $Loss_{Curve} \gets backward(P_{Curve}, label)$
\State $Loss_{Random} \gets backward(P_{Random}, label)$

\EndFor

\EndFor

\EndFor

\end{algorithmic}
\end{algorithm*}

\begin{table*}
  \centering
  \begin{tabular}{c | c | c}
    \hline
    Method & OA ↑ & mCE \\
    \hline
    DGCNN & 93.0\% & 0.669\\
    DGCNN + BatchNorm & 92.7\% & 0.527\\
    \hline
    DGCNN (W.M) & 93.2\% & 0.590\\
    DGCNN (W.M) + BatchNorm & 92.0\% & 0.512\\
    \hline
  \end{tabular}
  \caption{{\bf BatchNorm at test time.} Whereas overall accuracy is slightly degraded, OOD robustness is significantly increased, yielding lower $mCE$. This violates standard OOD assumptions but may be useful in some scenarios.}
  \label{table:batchnorm}
\end{table*}



    

\end{document}